\documentclass{article}

\usepackage[utf8]{inputenc}
\usepackage[T1]{fontenc}
\usepackage[preprint]{neurips_2026}

\usepackage{microtype}
\usepackage{graphicx}
\usepackage{subcaption}
\usepackage{booktabs}
\usepackage{multirow}
\usepackage{hyperref}
\usepackage{url}
\usepackage{amsmath}
\usepackage{amssymb}
\usepackage{mathtools}
\usepackage{amsfonts}
\usepackage{nicefrac}
\usepackage{xcolor}
\usepackage[capitalize,noabbrev]{cleveref}
\usepackage{placeins}

\newcommand{\cg}[1]{}
\newcommand{\mm}[1]{}

\title{Emergent Semantic Role Understanding \\in Language Models}

\author{%
  Carla Griffiths\\
  University College London\\
  \And
  Mirco Musolesi\\
  University College London\\
  University of Bologna
}

\begin{document}

\maketitle

\begin{abstract}
Understanding how linguistic structure emerges in language models is central to interpreting what these systems learn from data and how much supervision they truly require. In particular, semantic role understanding (``who did what to whom’’) is a core component of meaning representation, yet it remains unclear whether it arises from pre-training alone or depends on task-specific fine-tuning.

We study whether semantic role understanding emerges during language model pre-training or requires task-specific fine-tuning. We freeze decoder-only transformers and train linear probes to extract semantic roles, using performance to infer whether role information is already encoded in pre-training or learned during adaptation. Across model scales, we find that frozen representations contain substantial semantic role information, with performance improving but not fully matching fine-tuned models. This indicates partial but incomplete emergence from pre-training alone. We show that semantic role structure emerges from language modeling objectives, but its internal implementation shifts toward more distributed representations as model scale increases.
\end{abstract}

\section{Introduction}
\label{sec:introduction}

Understanding the emergence of linguistic structure in  Large Language Models (LLMs) is essential for characterizing what they learn from data and the extent to which supervision is required. In particular, \textit{semantic role understanding} (``who did what to whom’’) \cite{gildea2002automatic} is a core component of meaning representation, yet it remains unclear whether it arises from pre-training alone or depends on task-specific fine-tuning. Indeed, LLMs trained on next-token prediction have demonstrated capabilities that arguably extend beyond simple text completion \citep{brown2020language,radford2019language}. Building on the transformer architecture \citep{vaswani2017attention} and pre-training paradigms pioneered by BERT \citep{devlin2019bert} and ELMo \citep{peters2018deep}, modern LLMs learn rich contextual representations that generalize across tasks \citep{howard2018universal,ruder2019transfer}. This success has prompted extensive investigation into what linguistic knowledge these models acquire \citep{rogers2020primer,jawahar2019bert,liu2019linguistic}. 


Our central hypothesis is that \textit{semantic role understanding emerges from pre-training alone}, even at small model scales, and that this emergence can be detected and characterized through probing and targeted ablations. Semantic role labeling (SRL) is the task of identifying the predicate-argument structure of sentences, determining ``who did what to whom, when, where, and how'' \citep{palmer2010semantic}. For example, given the sentence ``The dog took a long walk in the park,'' QA-SRL generates questions like ``Who took a walk?'' (Agent: the dog), ``What did someone take?'' (Patient: a long walk), and ``Where did someone take a walk?'' (Location: the park). This formulation is particularly suitable for probing language models, as it requires no explicit linguistic annotation schema and naturally maps to extractive question answering. 

How can we tell if a model learned semantic roles during pre-training rather than during task-specific training? We freeze the pre-trained model and train only a simple linear layer to extract answers. If this frozen model performs well, semantic role information was encoded during pre-training; if the frozen model performs poorly, the capability must be learned during fine-tuning. In particular, our experimental design addresses three core questions:
\begin{enumerate}
    \item \textit{Does emergence of semantic role understanding occur?} Do frozen model representations contain sufficient information for semantic role identification?
    \item \textit{Is emergence of semantic role understanding scale-dependent?} At what model scale (if any) does the gap between frozen-probe and full-fine-tuning performance narrow significantly?
    \item \textit{How is semantic role information organized?} Can we identify how semantic roles are encoded inside the model and the intrinsic neural mechanisms? Do these representations form a clear, interpretable structure?
\end{enumerate}

Beyond linear probing, we perform mechanistic analysis to understand how semantic roles are represented. Using PCA and t-SNE \citep{vandermaaten2008tsne} visualization of layer activations, following similar approaches in LLM interpretability \citep{gurnee2023language}, we find that semantic roles form distinct clusters in representation space (Figure~\ref{fig:role-representations}), with separation increasing in deeper layers. We identify individual neurons in the feed-forward layers that selectively activate for specific semantic roles (e.g., ``\textit{Agent} neurons'', ``\textit{Location} neurons'') and causally validate their importance through targeted ablation experiments. These role-selective neurons exhibit functional co-activation patterns, with within-role correlations exceeding those across roles. This separation generally increases with layer depth.

We train decoder-only transformer models from scratch on WikiText-103 \citep{merity2016pointer}, spanning four scales from 0.4M to 57M transformer parameters. We evaluate using QA-SRL \citep{he2015question}, a task that reformulates semantic role labeling as question answering (e.g., asking ``Who took a walk?'' to identify the Agent role). Our results characterize the emergence threshold for semantic role understanding and contribute to the broader understanding of what linguistic capabilities arise from language modeling objectives alone.


This work makes the following contributions:
\begin{itemize}
    \item We present a systematic methodology for testing emergence of semantic role understanding using linear probing on frozen LLM representations. We provide empirical evidence characterizing the relationship between model scale and the frozen-probe vs. full-fine-tuning performance gap on QA-SRL.
    \item We introduce a temporal analysis framework using Centered Kernel Alignment (CKA) to track how representations evolve during fine-tuning relative to the pre-trained (null) model.
    \item We identify role-selective neurons through PCA/t-SNE visualization and neuron ablation, discovering that semantic roles cluster in representation space and that role-selective neurons show increasing co-activation separation with layer depth.
    \item We conduct role-conditioned neuron ablations revealing scale-dependent neural reorganization: \textit{Agent}-selective neurons transition from causally important in small models to interfering in large models, while \textit{Time}-selective neurons remain consistently causal across scales.
\end{itemize}
Code and trained models with checkpoint weights are available from the authors upon request.

\section{Related Work}
\label{sec:background}

\paragraph{Emergence in Language Models.}
\textit{Emergent abilities} in language models refer to capabilities that appear suddenly as models scale \citep{wei2022emergent}. Scaling laws predict smooth improvements in perplexity \citep{kaplan2020scaling,hoffmann2022training}, but task performance can exhibit sharp transitions. However, recent work questions whether emergence is a genuine phase transition or a measurement artifact: \citet{schaeffer2023emergent} argue that nonlinear metrics like exact match create the appearance of sudden capability acquisition, while \citet{snell_predicting_2024} show that fine-tuning can unlock capabilities present but inaccessible in smaller models. These findings suggest that ``emergence'' may reflect our evaluation methods rather than underlying model organization. Our work addresses this limitation by using continuous F1 metrics and linear probing, which provide a more fine-grained view of capability development. Rather than asking \textit{when} semantic role understanding appears, we characterize \textit{how much} is present at each scale, distinguishing between information encoded during pre-training versus information created during fine-tuning.
\cg{added our methodology vs prior work 25/01.}

\paragraph{Probing Classifiers.}
Probing classifiers analyze what information is encoded in neural representations \citep{belinkov2022probing}. The BERTology literature \citep{rogers2020primer,jawahar2019bert,clark2019what,liu2019linguistic,vulic2020probing,peters2018dissecting} has extensively applied probing, finding that lower layers encode syntax while higher layers capture semantics. Linear probes are preferred to constrain capacity \citep{hewitt2019designing}. Our methodology extends probing by comparing frozen performance to full fine-tuning, measuring how much information is already present versus must be learned.

\paragraph{Semantic Role Labeling.}
SRL has evolved from feature-based classifiers \citep{gildea2002automatic,pradhan2005semantic} to neural approaches \citep{zhou2015end,he2017deep,marcheggiani2017encoding,roth2016neural,strubell2018linguistically}, with pre-trained models providing strong features \citep{shi2019simple}. We use QA-SRL \citep{he2015question,fitzgerald2018large}, which reformulates SRL as question answering without requiring explicit PropBank schemas, making it suitable for probing emergence from language exposure alone.

\paragraph{Linguistic Structure in LLMs.}
Structural probes find syntactic tree structure in LLM representations \citep{hewitt2019structural}, while layer-wise analysis shows that lower layers encode surface features and higher layers capture abstract semantics \citep{tenney2019bert,peters2018dissecting,liu2019linguistic}. Building on static \citep{mikolov2013distributed,pennington2014glove} and contextualized embeddings \citep{peters2018deep,devlin2019bert}, our work examines semantic \textit{role} structure specifically, using ablations to establish causal relationships between neurons and semantic functions.

\paragraph{Mechanistic Interpretability.}
Recent work reverse-engineers neural network algorithms: feed-forward layers as key-value memories \citep{geva2021transformer}, circuit-level task analysis \citep{wang2022interpretability,conmy2023automated,olsson2022context}, and sparse autoencoders for monosemantic features \citep{elhage2022toy,bricken2023monosemanticity,cunningham2023sparse,templeton2024scaling}. Our work differs by training models from scratch for controlled scale manipulation, focusing on QA-SRL as a well-defined semantic task, and combining probing with neuron ablation to test causal necessity.

\section{Methods}
\label{sec:methods}

\subsection{Model Architecture}
We train decoder-only transformer language models following the GPT-2 \& GPT-3 architecture \citep{radford2019language}. All models use learned positional embeddings, pre-layer normalization, GELU activations, and a feed-forward expansion factor of 4. We use the \texttt{bert-base-uncased} tokenizer ($\sim$30K vocabulary), held fixed across all four scales following standard practice in scaling-law studies \citep{kaplan2020scaling,hoffmann2022training}. Scaling vocabulary with model size would confound transformer capacity with tokenization granularity; under our design, the same input distribution is presented to all four transformers, so the only varying quantity is the transformer's depth and width. We tie token embedding weights to the LM head during pre-training. We train four model configurations spanning approximately three orders of magnitude in parameter count (Table~\ref{tab:model-configs}).

\begin{table}[h]
\caption{Model configurations and parameter counts. Total parameters include embeddings (token + positional); transformer parameters exclude embeddings to enable comparison across vocabulary sizes.}
\label{tab:model-configs}
\centering
\small
\begin{tabular}{lccccc}
\toprule
Config & Layers & Hidden & Heads & Transformer & Total \\
\midrule
Tiny   & 2 & 128 & 2  & 0.4M  & 4.4M  \\
Small  & 4 & 256 & 4  & 3.2M  & 11.1M \\
Base   & 6 & 512 & 8  & 18.9M & 34.8M \\
Medium & 8 & 768 & 12 & 56.7M & 80.5M \\
\bottomrule
\end{tabular}
\end{table}

\subsection{Pre-training}

All models are pre-trained on WikiText-103 \citep{merity2016pointer} using the standard causal language modeling objective. WikiText-103 contains approximately 103 million tokens of Wikipedia articles, providing exposure to diverse factual content and linguistic structures without explicit QA formatting.

The pre-training objective is next-token prediction with cross-entropy loss:
\begin{equation}
\mathcal{L}_{\text{pretrain}} = -\sum_{t=1}^{T} \log P(x_t | x_{<t}; \theta)
\end{equation}
where $x_t$ is the token at position $t$ and $\theta$ represents model parameters.

\subsection{QA-SRL Fine-tuning Task}

After pre-training, models are fine-tuned (or probed) on QA-SRL \citep{he2015question}. Each example consists of a sentence, a predicate within that sentence, a question about that predicate's semantic role, and an answer span. During preprocessing, we skip examples where the answer text appears multiple times in the context (ambiguous spans), filtering approximately 9\% of examples. The model receives the concatenated input:
\begin{center}
\texttt{[question] [SEP] [sentence]}
\end{center}
and must predict start and end positions for the answer span within the sentence. Critically, we maintain causal attention during fine-tuning (matching pre-training), ensuring that any performance gap between frozen and fine-tuned models reflects differences in learned representations rather than attention pattern mismatch. With the question preceding the sentence, the model can attend to the full question when processing each sentence token.

We add a QA head consisting of two linear projections from the hidden dimension to scalar logits for start and end positions:
\begin{align}
\mathbf{s} &= \mathbf{H} \mathbf{w}_{\text{start}} \\
\mathbf{e} &= \mathbf{H} \mathbf{w}_{\text{end}}
\end{align}
where $\mathbf{H} \in \mathbb{R}^{L \times d}$ is the sequence of hidden states and $\mathbf{w}_{\text{start}}, \mathbf{w}_{\text{end}} \in \mathbb{R}^d$ are learned projection vectors.

\subsection{Linear Probing Protocol}

\paragraph{Overview} Our protocol distinguishes two related claims that prior probing work has often conflated. Information is \textit{decodable} when a linear probe extracts it from a frozen pre-trained model, establishing the information is present somewhere in the model's parameters and inputs together. It is \textit{emergent} from pre-training when the information lives in the pre-trained transformer weights, separately from trainable embeddings or fine-tuned layers, evidenced by a gap over an architecture-matched random baseline and corroborated by per-layer probes on pure pre-trained representations with zero embedding adaptation (Section~\ref{sec:layerwise}). Decodability is the weaker, established property; emergence under controlled conditions is the stronger claim our design tests. Note that the token embedding is a position-independent lookup table (identical outputs for ``the dog bit the child'' and ``the child bit the dog''), so role resolution under the frozen probe must come from the frozen transformer's cross-position attention rather than from embedding adaptation alone. We use two fine-tuning regimes described below.

\paragraph{Full Fine-tuning and Frozen Linear Probe.} In \textit{full fine-tuning}
, all model parameters (transformer layers and QA head) are updated during QA-SRL training. This establishes an upper bound on task performance for each model scale. In the case of the \textit{frozen linear probe}, all transformer block parameters and positional embeddings are frozen; only the token embeddings and QA head ($\mathbf{w}_{\text{start}}, \mathbf{w}_{\text{end}}$) are trained. This tests whether the answer span information is encoded in the frozen transformer representations. We deliberately keep the token embedding matrix trainable rather than frozen: the input format \texttt{[question] [SEP] [sentence]} relies on \texttt{[SEP]} as a structural separator, and \texttt{[SEP]} is a BERT-tokenizer artifact that was never seen in a separator role during causal language-model pre-training, so a fully frozen embedding matrix would prevent the model from parsing the input format and confound representation quality with input-format adaptation rather than test it. Embedding updates can still propagate through the frozen layers, creating a residual ``adaptation effect'' that we analyze separately via per-layer probes with zero embedding adaptation (Section~\ref{sec:layerwise}) and a stricter frozen-embedding (SEP-only) ablation in Appendix~\ref{app:sep-only}, where only \texttt{[SEP]} and the QA head update. The gap between full fine-tuning and frozen probe performance quantifies how much semantic role information must be learned during fine-tuning versus how much is already present from pre-training.

\paragraph{Random Baseline.} To establish a performance floor, we also evaluate a \textit{random baseline}: a randomly initialized (untrained) model whose token embeddings and QA head are trained on QA-SRL while the transformer blocks remain at random initialization. This is a capacity-matched comparator for the frozen probe (same trainable parameters, only the source of the transformer weights differs), which isolates the contribution of the pre-trained transformer from any contribution of trainable embeddings.

\paragraph{Normalized Emergence Score.} We define a normalized emergence score that accounts for the random baseline floor:
\begin{equation}
\text{Emergence}_{\text{norm}} = \frac{F1_{\text{frozen}} - F1_{\text{random}}}{F1_{\text{full}} - F1_{\text{random}}}.
\end{equation}
A score near 1.0 indicates strong emergence (frozen probe captures most of what full fine-tuning achieves above chance), while a score near 0.0 indicates the frozen probe performs no better than random initialization. We also report the raw gap $\Delta_{\text{emergence}} = F1_{\text{full}} - F1_{\text{frozen}}$ (while less rigorous, the raw score is more intuitive).

\subsection{Temporal CKA Analysis}
\label{sec:temporal-cka}

To understand how representations evolve during fine-tuning, we employ Centered Kernel Alignment (CKA) \citep{kornblith2019similarity} to measure representational similarity across training checkpoints. For activation matrices $\mathbf{X}, \mathbf{Y} \in \mathbb{R}^{n \times d}$, linear CKA is:
\begin{equation}
\text{CKA}(\mathbf{X}, \mathbf{Y}) = \frac{\|\mathbf{Y}^\top \mathbf{X}\|_F^2}{\|\mathbf{X}^\top \mathbf{X}\|_F \|\mathbf{Y}^\top \mathbf{Y}\|_F}.
\end{equation}
We save checkpoints at each epoch and compute CKA between each fine-tuning checkpoint and the pre-trained (\textit{null}) model. A decrease in $\text{CKA}_{\text{null}}$ indicates representational drift from pre-training during fine-tuning.

\section{Results}
\label{sec:results}

\subsection{Overview}
We present results across four model scales: Tiny (0.4M), Small (3.2M), Base (18.9M), and Medium (56.7M transformer parameters). We report the F1 Score on the QA-SRL validation step, defined as the token-level overlap between predicted and gold answer spans, computed as the harmonic mean of precision and recall.
In the evaluation, we apply context masking to constrain predictions to the sentence region only, preventing the model from predicting spans within the question or special tokens. 
Table~\ref{tab:results} summarizes performance across three experimental conditions for each model size: random baseline (untrained LLM with trained QA head), frozen linear probe (pre-trained LLM frozen, only QA head trained), and full fine-tuning (all parameters updated).

\begin{table}[t]
\caption{Results on QA-SRL semantic role labeling across model scales. All models show positive emergence, with frozen pre-trained representations encoding semantic role information accessible to linear probes. F1 (\%) shown as mean$\pm$std over 5 random seeds; emergence score $E = \Delta_{\text{pre}} / (\Delta_{\text{pre}} + \Delta_{\text{ft}})$.}
\label{tab:results}
\centering
\small
\begin{tabular}{lcccc}
\toprule
& Tiny (0.4M) & Small (3.2M) & Base (18.9M) & Medium (56.7M) \\
\midrule
Random baseline       & 24.2$\pm$0.6 & 25.3$\pm$0.1 & 25.5$\pm$0.2 & 26.1$\pm$0.5 \\
Frozen (linear probe) & 36.4$\pm$0.8 & 39.6$\pm$0.7 & 43.6$\pm$0.5 & 44.7$\pm$0.5 \\
Full fine-tuning      & 44.6$\pm$0.4 & 53.3$\pm$0.7 & 63.8$\pm$0.4 & 65.8$\pm$0.7 \\
\midrule
Emergence $E$         & 0.60$\pm$0.06 & 0.51$\pm$0.03 & 0.47$\pm$0.02 & 0.47$\pm$0.02 \\
\bottomrule
\end{tabular}
\end{table}

\begin{figure}[t]
\centering
\begin{subfigure}[b]{0.62\linewidth}
\includegraphics[width=\linewidth]{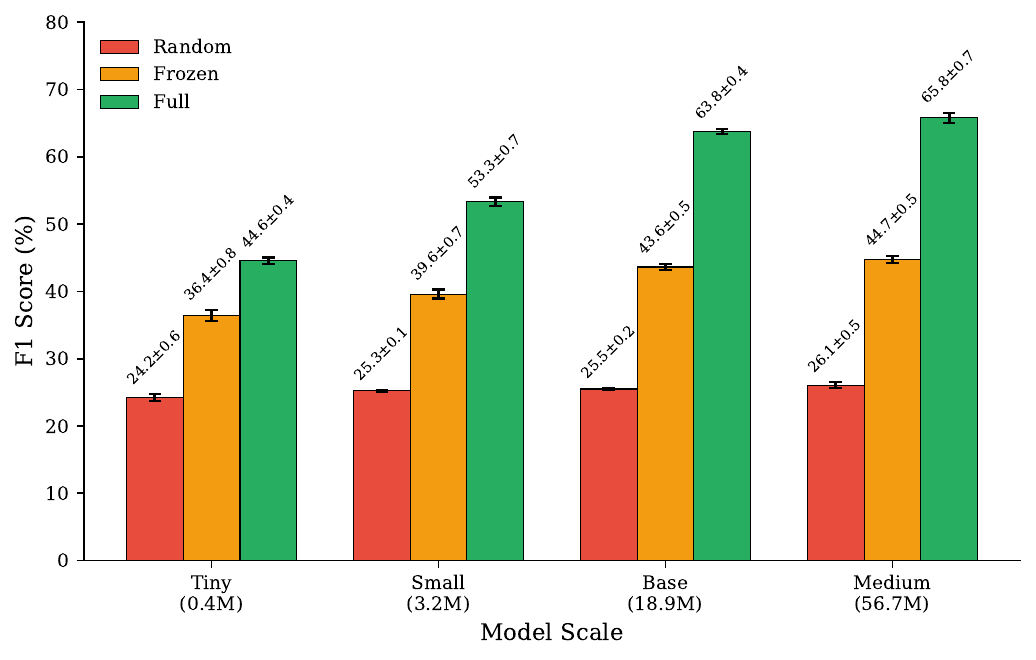}
\caption{F1 across conditions and scales.}
\label{fig:results-comparison}
\end{subfigure}
\hfill
\begin{subfigure}[b]{0.34\linewidth}
\includegraphics[width=\linewidth]{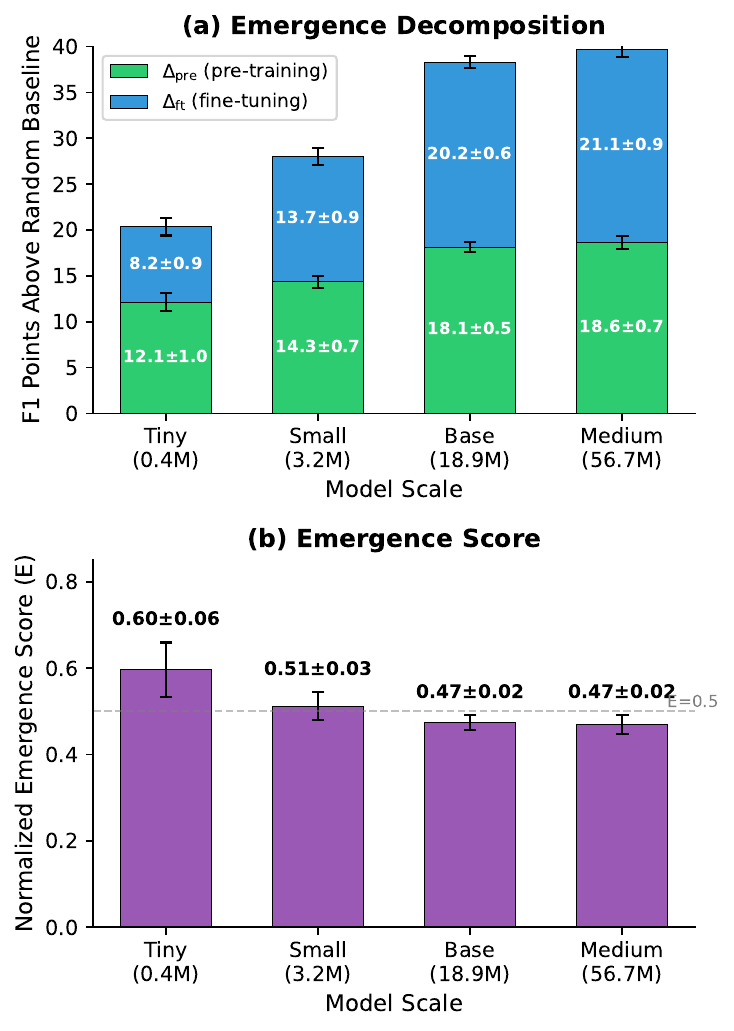}
\caption{Emergence decomposition.}
\label{fig:emergence-decomp}
\end{subfigure}
\caption{Frozen probes extract semantic role information without any model adaptation. (a) We compare three conditions: \textit{Random baseline} (untrained model + QA head only), \textit{Frozen probe} (pre-trained model frozen, only QA head trained), and \textit{Full fine-tuning} (all parameters updated). The frozen probe substantially outperforms random at all scales. (b) Decomposing performance into $\Delta_{\text{pre}} = F1_{\text{frozen}} - F1_{\text{random}}$ (pre-training above chance) and $\Delta_{\text{ft}} = F1_{\text{full}} - F1_{\text{frozen}}$ (fine-tuning's additional contribution): pre-training contributes 12--19 F1 points across scales while fine-tuning's contribution grows with scale (8--21 points). The normalized emergence score $E = \Delta_{\text{pre}} / (\Delta_{\text{pre}} + \Delta_{\text{ft}})$ decreases with scale because fine-tuning becomes increasingly effective relative to a relatively constant pre-training contribution. Values: mean $\pm$ std over 5 random seeds.}
\label{fig:results}
\end{figure}

\subsection{Analysis of the Emergence of Semantic Role Understanding}

All models show positive emergence, with frozen probes consistently outperforming random baselines (Figure~\ref{fig:emergence-decomp}). To understand what drives the emergence score, we decompose it into two components:
\begin{itemize}
    \item $\Delta_{\text{pre}} = F1_{\text{frozen}} - F1_{\text{random}}$: pre-training's contribution above chance;
    \item $\Delta_{\text{ft}} = F1_{\text{full}} - F1_{\text{frozen}}$: fine-tuning's additional contribution.
\end{itemize}
The normalized emergence score is defined as $E = \Delta_{\text{pre}} / (\Delta_{\text{pre}} + \Delta_{\text{ft}})$.
Examining these components reveals that pre-training's contribution ($\Delta_{\text{pre}}$) ranges from 12.1 to 18.6 F1 points (12.1$\pm$1.0, 14.3$\pm$0.7, 18.1$\pm$0.5, 18.6$\pm$0.7 for Tiny through Medium), showing consistent gains across all scales. In contrast, fine-tuning's contribution ($\Delta_{\text{ft}}$) grows substantially with scale (8.2$\pm$0.9, 13.7$\pm$0.9, 20.2$\pm$0.6, 21.1$\pm$0.9 points). The normalized emergence score decreases with scale because fine-tuning becomes increasingly effective at larger scales while pre-training's contribution stays roughly constant. The score thus reflects the fraction of task performance that comes ``for free'' from pre-training, and this fraction shrinks as fine-tuning scales up.

These results show that semantic role information is decodable from frozen pre-trained representations across all scales, with a consistent 12--19 F1 gap above the architecture-matched random baseline. The frozen-probe condition, however, leaves the token embedding matrix trainable, conflating emergence in transformer weights with task-driven embedding adaptation; we disentangle the two next.

\subsection{Representational Changes During Fine-tuning}
\label{sec:temporal-cka-results}

\begin{figure}[t]
\centering
\begin{subfigure}[b]{0.48\linewidth}
\includegraphics[width=\linewidth]{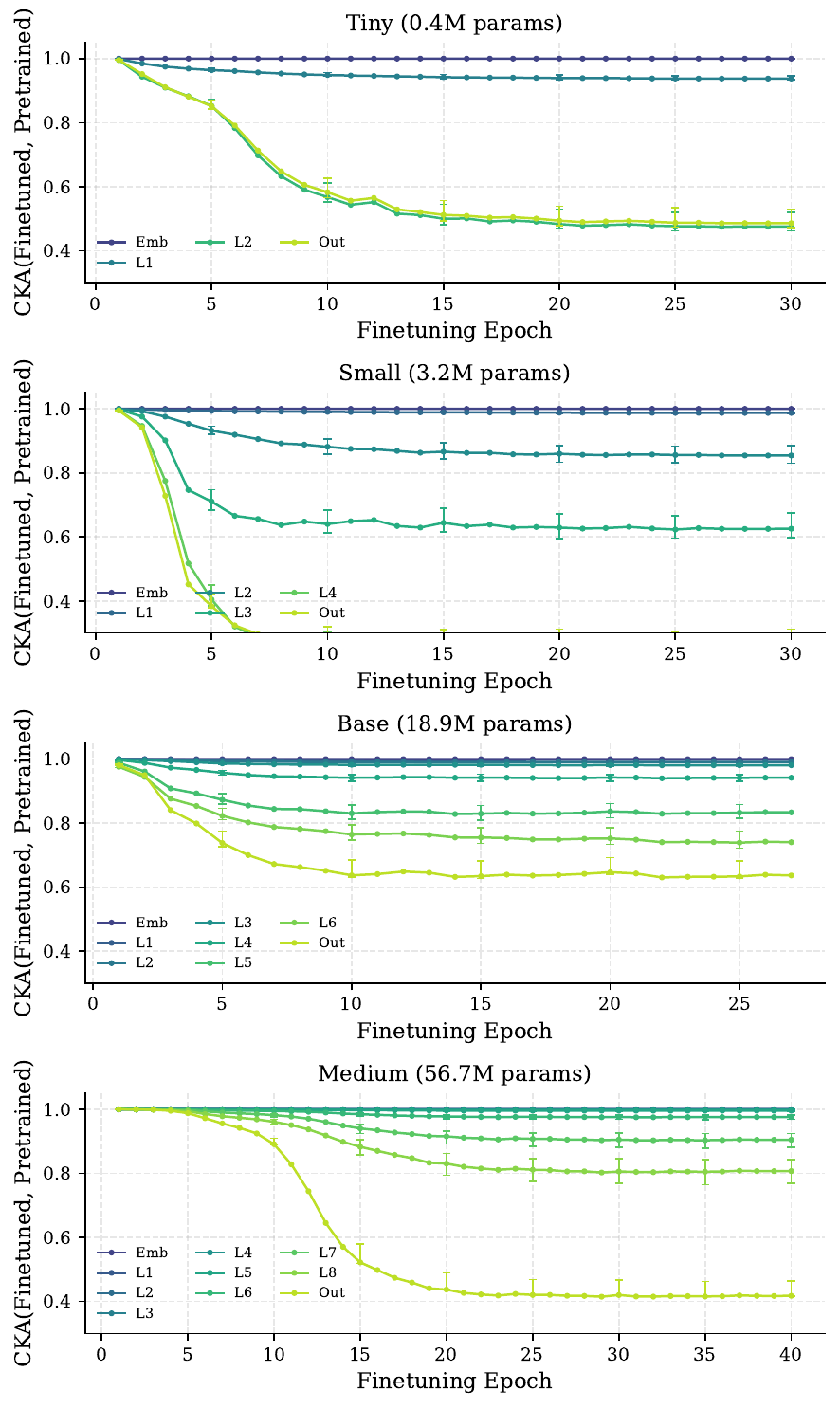}
\caption{Temporal CKA across layers.}
\label{fig:cka-temporal}
\end{subfigure}
\hfill
\begin{subfigure}[b]{0.48\linewidth}
\includegraphics[width=\linewidth]{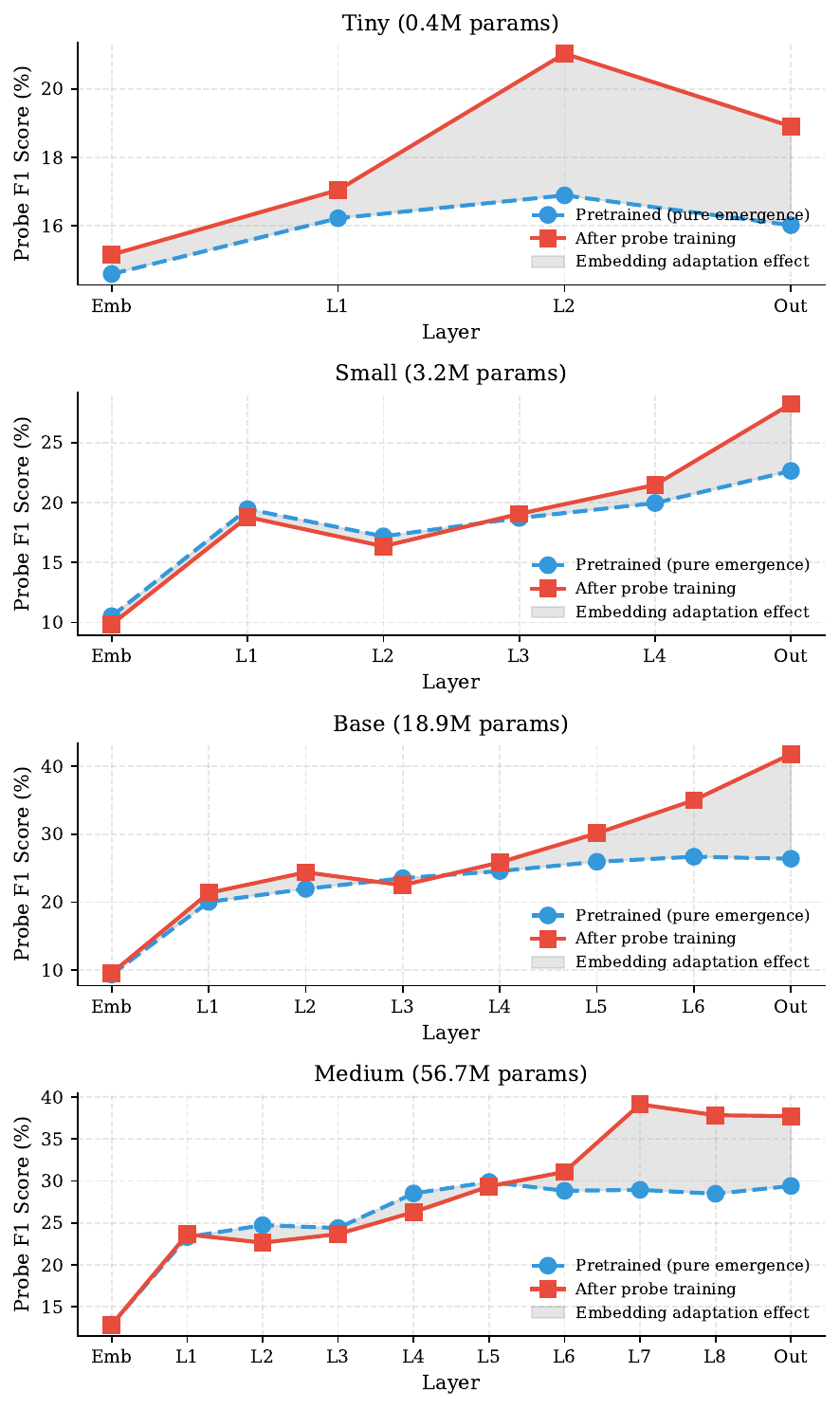}
\caption{Per-layer probing F1.}
\label{fig:layerwise-probing}
\end{subfigure}
\caption{Representational changes during fine-tuning, and isolation of pre-training from adaptation. (a) Centered Kernel Alignment (CKA) between each fine-tuning epoch and the original pre-trained model: deeper layers (darker lines) diverge most, dropping to CKA $\approx 0.25$ to $0.49$, while early layers remain stable. (b) Per-layer linear probes comparing \textit{Pure pre-trained} (blue dashed; before any SRL training) and \textit{After probe training} (red solid). Gray region marks adaptation effect. Probes are trained on 500 samples and measure layer-specific linear separability, so absolute values differ from Table~\ref{tab:results}. F1 on pure pre-trained representations ranges from 10 to 26\% across scales, isolating information present in fixed transformer weights.}
\label{fig:representations}
\end{figure}

To understand how representations evolve during fine-tuning, we perform Centered Kernel Alignment (CKA) analysis comparing each fine-tuning epoch against the pre-trained model. CKA analysis shows that full fine-tuning substantially reorganizes representations, with deeper layers diverging most from pre-training (CKA dropping to 0.25 to 0.49; Figure~\ref{fig:cka-temporal}). This raises a key question: is semantic role information \textit{created} during fine-tuning, or already present in pre-trained representations?

\subsection{Isolating Emergence from Adaptation}
\label{sec:layerwise}

To determine whether semantic role information is encoded during pre-training (emergence) versus created during fine-tuning (adaptation), we train linear probes on pre-trained representations before any task-specific training. We compare this to probes trained after frozen probe training.
Semantic role information is present in pure pre-trained representations: linear probes achieve 10 to 26\% F1 across layers and scales before any SRL training (Figure~\ref{fig:layerwise-probing}). This confirms that emergence occurs even at small scales; predicate-argument structure is encoded during pre-training rather than created during task-specific adaptation. As a stricter control, a frozen-embedding (SEP-only) ablation isolates the residual transformer contribution and is statistically significant at our two largest scales ($p=0.004$, $p=0.002$; full results in Appendix~\ref{app:sep-only}).

\subsection{Neuron-Level Dynamics}
\label{sec:neurons}

\paragraph{Overview.} We identify role-selective neurons using PCA. First, we run PCA on feed-forward activations across all validation samples, then test whether principal components separate semantic roles using ANOVA F-statistics. For each discriminative principal component ($p < 0.0001$), we extract the top 5 neurons with the highest absolute loadings, selecting up to 20 role-selective neurons per layer across all discriminative principal components. 
We refer to neurons selective for specific PropBank roles: \textit{Agent} neurons (selective for ARG0, the actor/doer of an action), \textit{Time} neurons (selective for ARGM-TMP, temporal modifiers like ``yesterday'' or ``during the meeting''), \textit{Location} neurons (ARGM-LOC), and \textit{Manner} neurons (ARGM-MNR) (PropBank definitions in Section \ref{app:propbank}).
\cg{added italics; kept ``Manner'' as standard PropBank terminology.}

\paragraph{PCA and t-SNE Visualizations.} PCA and t-SNE visualizations of final-layer feed-forward activations (Appendix~\ref{app:representations}, Figure~\ref{fig:role-representations}) reveal that activations show partial organization by PropBank semantic role, with ARG0-Agent showing the clearest separation; the first two principal components explain only 14.8\% of variance, indicating that semantic roles are encoded in a higher-dimensional subspace not fully captured by linear projections. Named-role-selective neurons (\textit{Agent}, \textit{Location}, \textit{Manner}) concentrate in early layers (L0--L1; see Table~\ref{tab:selective-neurons} and Figure~\ref{fig:neuron-selectivity} in Appendix~\ref{app:neurons}), suggesting role encoding begins early in the network. To causally validate neuron importance, we set the weights of PCA-identified neurons to 0 in the \textit{pretrained} model, then finetune on QA-SRL and evaluate. This tests whether specific neurons are necessary (not just correlated) for learning semantic roles. We perform role-conditioned neuron ablation: ablating role $X$'s neurons and measuring role $X$'s specific F1 (Table~\ref{tab:role-conditioned-ablation}, for full results see Table~\ref{tab:neuron-ablation-full}).

\begin{figure}[t]
\centering
\includegraphics[width=0.58\linewidth]{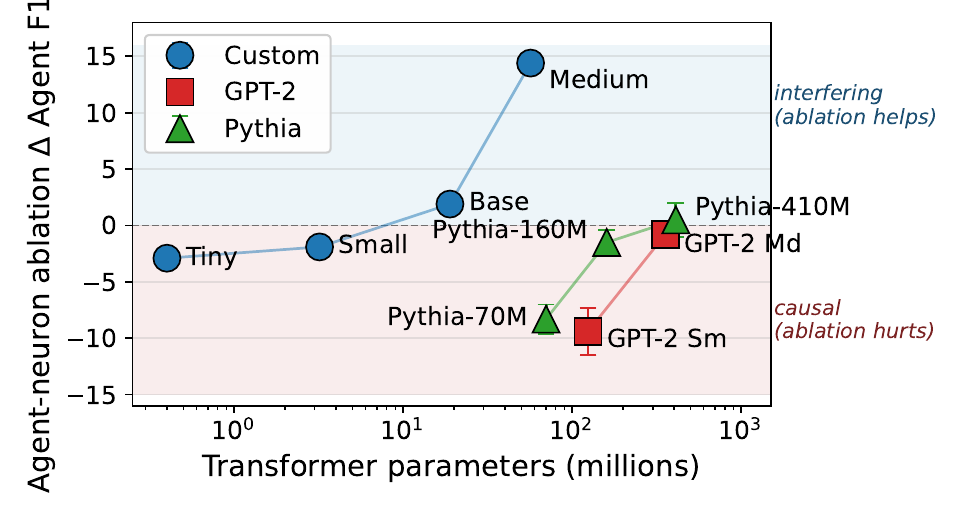}
\caption{Ablation $\Delta$\textit{Agent} F1 (role-specific, matching Table~\ref{tab:role-conditioned-ablation}; not overall F1) across 9 parameter configurations. The custom-model trajectory (blue) crosses zero between Small and Base; GPT-2 and Pythia show the same pattern. Error bars: $\pm$ std over 5 seeds.}
\label{fig:sign-reversal}
\end{figure}

\begin{table}[t]
\caption{Role-conditioned neuron ablation. We ablate each role's selective neurons and measure \textit{that role's} F1 specifically (i.e., F1 computed only on validation examples whose gold answer carries that role label, not overall task F1 from Table~\ref{tab:results}). Negative $\Delta$ confirms causal importance for that semantic role; positive $\Delta$ in larger models indicates interference. F1 baselines for each role differ from overall F1 because role-specific subsets are smaller relative to the whole population.}
\label{tab:role-conditioned-ablation}
\begin{center}
\begin{tabular}{lcccccccc}
\toprule
& \multicolumn{2}{c}{Tiny} & \multicolumn{2}{c}{Small} & \multicolumn{2}{c}{Base} & \multicolumn{2}{c}{Medium} \\
Role Ablated & F1 & $\Delta$ & F1 & $\Delta$ & F1 & $\Delta$ & F1 & $\Delta$ \\
\midrule
\textit{Agent} $\rightarrow$ \textit{Agent} F1 & 0.0\% & -2.9\% & 11.5\% & -1.9\% & 14.4\% & +1.9\% & 24.0\% & +14.4\%$^\dagger$ \\
\textit{Time} $\rightarrow$ \textit{Time} F1 & 2.4\% & -2.4\% & 9.8\% & -4.9\% & 22.0\% & -2.4\% & 7.3\% & -4.9\% \\
\textit{Location} $\rightarrow$ \textit{Location} F1 & 0.0\% & 0.0\% & 0.0\% & 0.0\% & 3.6\% & 0.0\% & 3.6\% & 0.0\% \\
\textit{Manner} $\rightarrow$ \textit{Manner} F1 & 3.8\% & +3.8\% & 0.0\% & 0.0\% & 3.8\% & 0.0\% & 3.8\% & +3.8\% \\
\bottomrule
\multicolumn{9}{l}{\footnotesize $^\dagger$Positive $\Delta$ suggests interference or redundant representations in larger models.}
\end{tabular}
\end{center}
\end{table}
\paragraph{Role-conditioned Analysis.} The role-conditioned results reveal scale-dependent neural organization. \textit{Time} neurons show consistent causal importance across all scales, with ablations reducing \textit{Time} F1 by $-2.4\%$ to $-4.9\%$. Smaller models (Tiny: $-2.9\%$, Small: $-1.9\%$) show clear dependence on \textit{Agent}-selective neurons, while larger models show improved performance when these neurons are ablated (Base: $+1.9\%$, Medium: $+14.4\%$). This reversal suggests that larger models develop redundant or interfering representations, and that ablating putatively \textit{Agent}-selective neurons removes this interference. \textit{Location} and \textit{Manner} neurons show minimal or no causal effects across scales ($\Delta \approx 0$), suggesting these roles lack specialized neurons or are represented distributionally. The consistent \textit{Time} neuron effects, combined with scale-dependent \textit{Agent} effects, demonstrate that role-selective neurons exist but reorganize with capacity. The Agent-neuron sign attenuation replicates on GPT-2 (124M, 355M; Appendix~\ref{app:gpt2}) and Pythia (70M, 160M, 410M; Appendix~\ref{app:pythia}) with random-neuron controls confirming role-specificity, and component-level circuit analysis across all six models shows the same concentrated-to-distributed transition (Appendix~\ref{app:circuit}), Figure~\ref{fig:sign-reversal}).

\paragraph{Cross-role Interference and Co-activation.} Two distinct patterns of interference emerge at scale, both absent in small models. (i) \textit{Time}-neuron ablation in larger models reduces \textit{Time} F1 ($-4.9\%$) but improves overall F1 ($+1.0\%$): the role-specific neurons are causally helpful for their own role yet suppress more frequent roles, so removing them improves overall F1 via weighted averaging. (ii) Medium-scale \textit{Agent}-neuron ablation improves both \textit{Agent} F1 ($+14.4\%$) and overall F1 ($+5.8\%$): putatively \textit{Agent}-selective neurons have become net interfering even for their own role, consistent with redundant or miscoordinated representations at scale. Pairwise correlations between PCA-identified neurons show same-role neurons more correlated than different-role neurons ($\rho_{\text{within}} \approx 0.24$ vs.\ $\rho_{\text{across}} \approx 0.10$), consistent with distributed, overlapping representations. Per-role critical layers, overall ablation F1, and the power-law fit are in Appendix~\ref{app:neurons} and Appendix~\ref{app:ablation}.

\paragraph{Limitations.}\label{sec:limitations} \textit{Single-seed ablations:} role-conditioned and overall neuron-ablation results on custom models (Tables~\ref{tab:role-conditioned-ablation},~\ref{tab:neuron-ablation-full},~\ref{tab:role-layer-critical}) are point estimates from one fine-tuning run per (model, role) pair; the qualitative \textit{Agent}-neuron sign-attenuation pattern holds across 5-seed replications on GPT-2 and Pythia. \textit{Dataset quality:} the original QA-SRL dataset \citep{he2015question} achieves ${\sim}72.4\%$ recall against expert annotations \citep{fitzgerald2018large}, which may underestimate absolute F1; our analyses focus on the emergence \textit{gap}, where systematic annotation noise cancels. \textit{Data-constrained scaling:} WikiText-103's ${\sim}103$M tokens give a tokens-to-parameters ratio of ${\sim}1.8$ for Medium versus the Chinchilla-optimal ${\sim}20$ \citep{hoffmann2022training}; despite this, Medium achieves the highest fine-tuned F1 ($65.8\% \pm 0.7\%$), and reported emergence is conservative as Medium remains under-converged (Appendix~\ref{app:pretraining-emergence}). The Agent-neuron sign reversal replicates on fully-trained GPT-2 and Pythia (Appendices~\ref{app:gpt2}--\ref{app:pythia}). \textit{Model scale:} custom models reach 57M transformer parameters; GPT-2 (355M) and Pythia (410M) remain small relative to massive commercialized LLMs.

\section{Conclusion}
\label{sec:conclusion}
\paragraph{Summary.} 
We have studied whether semantic role understanding emerges during language model pre-training and presented an investigation of how semantic role understanding emerges in language models across four scales (0.4M to 57M transformer parameters). We have showed that pre-training contributes 12--19 F1 above architecture-matched random initialization across scales, with layer-wise probing on pure pre-trained representations confirming encoding of predicate-argument structure. Moreover, feed-forward neurons activate selectively for semantic roles, with co-activation separation generally increasing with layer depth. Role-conditioned ablations reveal consistent importance of time-related neurons across all scales
while neurons associated to the agent show scale-dependent effects: causal in smaller models 
but improving performance when ablated in larger models, 
suggesting neural reorganization as model size increases. Critically, cross-role interference emerges in larger models. 
This scale-dependent reorganization shows that larger models develop complex neuron interactions absent in smaller models.

\paragraph{Implications.} Our work has important implications for understanding language models. First, role-selective neurons demonstrate that emergent linguistic capabilities can partially localize, suggesting that interpretability methods can identify meaningful structure in emergent behaviors. Second, cross-role interference in larger models demonstrates that language models do not necessarily implement clean one-to-one mappings between individual neurons and specific functions. Third, the scale-dependent neural reorganization, where \textit{Agent} neurons transition from causally important to interfering, challenges the assumption that circuits identified in smaller models will retain their function at larger scales. Fourth, our frozen probing methodology suggests that semantic role understanding emerges from pre-training on natural language across model scales.

\begin{ack}
This work investigates the emergence of semantic role understanding through assessing what capabilities are learned during pre-training versus fine-tuning. We see no ethical issues with this work, as understanding how semantic understanding emerges helps demystify LLM capabilities. Our experiments required approximately 860 GPU hours for the custom models plus approximately 200 GPU hours for the GPT-2 and Pythia replications; we will release the full model weights with checkpoints to reduce redundant computation. We thank Modal for a GPU grant that enabled the majority of these experiments to be run at no cost.
\end{ack}

\FloatBarrier
\bibliographystyle{unsrtnat}
\bibliography{example_paper}

\newpage
\appendix

\section{Full Neuron Ablation Results}
\label{app:ablation}

Table~\ref{tab:neuron-ablation-full} shows overall F1 impact when ablating PCA-identified role-selective neurons. These results aggregate across all semantic roles in the validation set and reflect weighted averaging due to unbalanced role distributions (\textit{Theme}: 55.8\%, \textit{Agent}: 20.8\%, \textit{Time}: 8.2\%, \textit{Location}: 5.6\%, others: 9.6\%). For more interpretable causal validation, see Table~\ref{tab:role-conditioned-ablation}, which measures role-specific F1 impact.

\begin{table}[h]
\caption{Overall neuron ablation results across all semantic roles. We ablate PCA-identified role-selective neurons before fine-tuning and measure overall F1 impact. Baseline F1 scores in column headers (e.g., ``Tiny (2.2\%)'') show pre-ablation overall F1 \textit{on the role-conditioned evaluation subset used in Table~\ref{tab:role-conditioned-ablation}}, not overall F1 on the full validation set from Table~\ref{tab:results}. These overall F1 changes reflect weighted averaging across unbalanced role distributions and may not align with role-specific effects (see Section~\ref{sec:neurons} and Table~\ref{tab:role-conditioned-ablation}). Values are single-seed point estimates.}
\label{tab:neuron-ablation-full}
\begin{center}
\begin{small}
\resizebox{0.8\textwidth}{!}{%
\begin{tabular}{lcccccccc}
\toprule
& \multicolumn{2}{c}{Tiny (2.2\%)} & \multicolumn{2}{c}{Small (6.2\%)} & \multicolumn{2}{c}{Base (7.6\%)} & \multicolumn{2}{c}{Medium (5.4\%)} \\
Ablation & F1 & $\Delta$ & F1 & $\Delta$ & F1 & $\Delta$ & F1 & $\Delta$ \\
\midrule
\textit{Agent} & 1.2\% & -1.0\% & 5.0\% & -1.2\% & 7.4\% & -0.2\% & 11.2\% & +5.8\%$^\dagger$ \\
\textit{Location} & 2.2\% & 0.0\% & 5.0\% & -1.2\% & 8.0\% & +0.4\% & 6.6\% & +1.2\% \\
\textit{Time} & 1.2\% & -1.0\% & 6.0\% & -0.2\% & 7.4\% & -0.2\% & 6.4\% & +1.0\% \\
\textit{Manner} & 2.4\% & +0.2\% & 5.8\% & -0.4\% & 8.8\% & +1.2\% & 10.0\% & +4.6\%$^\dagger$ \\
\bottomrule
\multicolumn{9}{l}{\footnotesize $^\dagger$Positive $\Delta$ in larger models reflects cross-role interference (Section~\ref{sec:neurons}).}
\end{tabular}%
}
\end{small}
\end{center}
\end{table}

\section{Training Hyperparameters}
\label{app:hyperparameters}

All experiments were conducted on NVIDIA GPUs using Modal's cloud infrastructure. Models were implemented in PyTorch 2.0 with mixed-precision training (FP16) where applicable. We generated the analysis for Figures 3-8 locally using Metal Performance Shaders on an M4 MacBook Pro.

\subsection{Compute Resources}

\begin{table}[h]
\caption{Compute configuration and estimated training times per model. Base and Medium models were migrated to faster GPUs mid-training due to time constraints (Base: epoch 78, Medium: epoch 68).}
\label{tab:compute}
\begin{center}
\begin{tabular}{lccccc}
\toprule
Config & GPU & VRAM & Pretrain Time & Finetune Time & Total GPU Hours \\
\midrule
Tiny & A10G & 24GB & $\sim$48h & $\sim$1h & $\sim$49h \\
Small & A10G & 24GB & $\sim$72h & $\sim$1h & $\sim$73h \\
Base & A10G $\rightarrow$ H200 & 24/80GB & $\sim$336h & $\sim$4h & $\sim$340h \\
Medium & A100 $\rightarrow$ H200 & 80GB & $\sim$390h & $\sim$10h & $\sim$400h \\
\bottomrule
\end{tabular}
\end{center}
\end{table}

\subsection{Pre-training Hyperparameters}

\begin{table}[h]
\caption{Pre-training hyperparameters for each model configuration.}
\label{tab:pretrain-hparams}
\begin{center}
\begin{small}
\begin{tabular}{lcccc}
\toprule
Hyperparameter & Tiny & Small & Base & Medium \\
\midrule
Sequence Length & 512 & 512 & 512 & 512 \\
Batch Size & 64 & 48 & 32 & 96 \\
Gradient Accumulation & 1 & 1 & 1 & 1 \\
Learning Rate & $5 \times 10^{-4}$ & $2 \times 10^{-4}$ & $2 \times 10^{-4}$ & $1 \times 10^{-4}$ \\
Warmup Steps & 500 & 500 & 1000 & 2000 \\
Max Epochs & 200 & 200 & 200 & 200 \\
Early Stopping Patience & 20 & 20 & 20 & 20 \\
Dropout Rate & 0.05 & 0.1 & 0.1 & 0.1 \\
Samples per Epoch & 200K & 400K & 800K & 1.5M \\
\bottomrule
\end{tabular}
\end{small}
\end{center}
\end{table}

\subsection{Fine-tuning Hyperparameters}

All fine-tuning results are reported as mean$\pm$std over 5 random seeds (42, 1042, 2042, 3042, 4042).

\begin{table}[h]
\caption{QA-SRL fine-tuning hyperparameters for each model configuration.}
\label{tab:fine-tune-hparams}
\begin{center}
\begin{small}
\begin{tabular}{lcccc}
\toprule
Hyperparameter & Tiny & Small & Base & Medium \\
\midrule
Sequence Length & 512 & 512 & 512 & 512 \\
Batch Size & 64 & 48 & 32 & 32 \\
Learning Rate (Full Finetune) & $4 \times 10^{-5}$ & $2 \times 10^{-5}$ & $1 \times 10^{-5}$ & $1 \times 10^{-5}$ \\
Learning Rate (Frozen Probe) & $1 \times 10^{-3}$ & $1 \times 10^{-3}$ & $1 \times 10^{-3}$ & $1 \times 10^{-3}$ \\
Max Epochs & 30 & 30 & 30 & 30 \\
Early Stopping Patience & 8 & 8 & 8 & 8 \\
\bottomrule
\end{tabular}
\end{small}
\end{center}
\end{table}

\subsection{Optimizer Configuration}

All models use the AdamW optimizer \citep{loshchilov2017decoupled} with the following settings:
\begin{itemize}
    \item $\beta_1 = 0.9$, $\beta_2 = 0.999$.
    \item Weight decay: $0.01$.
    \item Learning rate schedule: Constant learning rate with early stopping.
    \item Gradient clipping: Max norm $1.0$.
    \item Label smoothing: $\alpha = 0.1$ (pre-training cross-entropy loss).
\end{itemize}

\subsection{Temporal CKA Analysis Configuration}

For the temporal CKA analysis described in Section~\ref{sec:temporal-cka}:
\begin{itemize}
    \item Probe set size: 500 examples from QA-SRL validation set.
    \item Checkpoint frequency: Every fine-tuning epoch.
    \item Activation pooling: Mean pooling over sequence length (excluding padding).
    \item CKA variant: Linear CKA with centered activations.
\end{itemize}

\section{Neuron-Level Analysis Details}
\label{app:neurons}

\begin{table}[h]
\caption{Critical layer for each role: layer whose ablation gives the largest \textit{role-specific} $\Delta$F1 (computed on the subset of validation examples carrying that role label, not overall F1). Values in parentheses are role-specific $\Delta$F1 for that layer. Single-seed point estimates.}
\label{tab:role-layer-critical}
\begin{center}
\begin{tabular}{@{}lcccc@{}}
\toprule
 & Tiny & Small & Base & Med \\
\midrule
LOC & L0 ($-$17\%) & L1 ($-$12\%) & L1 ($-$6\%) & L0 ($-$13\%) \\
MNR & L1 ($-$18\%) & L3 ($-$24\%) & L0 ($-$11\%) & L5 ($-$5\%) \\
ARG2 & L0 ($-$14\%) & L0 ($-$4\%) & L0 ($-$12\%) & L0 ($-$5\%) \\
TMP & L1 ($-$6\%) & L1 ($-$9\%) & L4 ($-$3\%) & L0 ($-$6\%) \\
\bottomrule
\end{tabular}
\end{center}
\end{table}

We identify neurons with selectivity scores $> 10\times$ the mean activation for specific roles:

\begin{table}[h]
\caption{Top role-selective neurons in the Small model. Selectivity is computed as the ratio of mean activation for the target role to mean activation across all roles.}
\label{tab:selective-neurons}
\begin{center}
\begin{tabular}{llcc}
\toprule
Role & Layer & Neuron Index & Selectivity \\
\midrule
\textit{Agent} & 0 & 41 & 40.7$\times$ \\
\textit{Agent} & 1 & 343 & 28.3$\times$ \\
\textit{Location} & 0 & 180 & 31.5$\times$ \\
\textit{Location} & 1 & 530 & 36.2$\times$ \\
\textit{Time} & 0 & 665 & 22.1$\times$ \\
\textit{Time} & 1 & 817 & 19.8$\times$ \\
\textit{Time} & 1 & 826 & 18.4$\times$ \\
\textit{Manner} & 0 & 412 & 15.2$\times$ \\
\textit{Manner} & 2 & 891 & 14.7$\times$ \\
\bottomrule
\end{tabular}
\end{center}
\end{table}

Table~\ref{tab:selective-neurons} lists the most selective neurons for each semantic role. This specialization emerges naturally from language model pre-training without explicit role supervision.

\begin{figure}[h]
\centering
\includegraphics[width=0.65\linewidth]{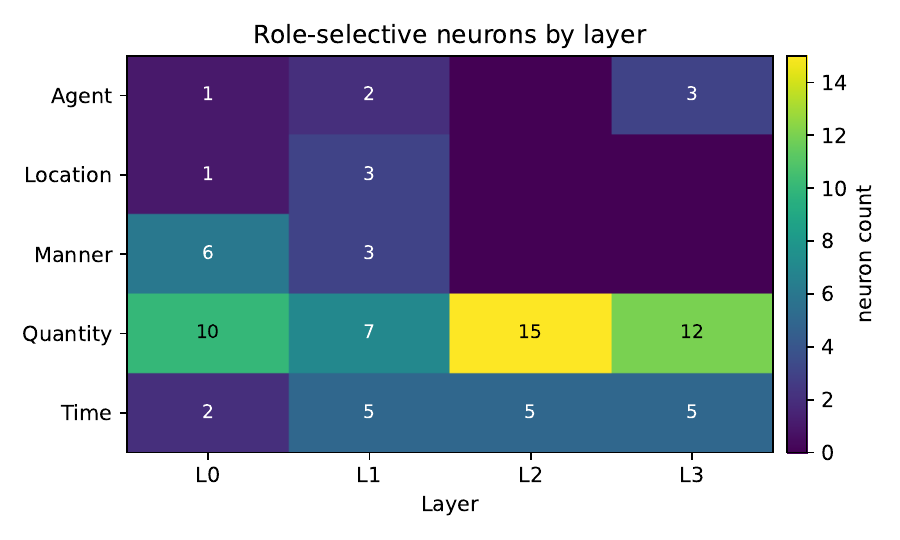}
\caption{Layer $\times$ role distribution of the top-20 PCA-identified selective neurons per layer in the Small (3.2M) model. Cell values are neuron counts. Named PropBank-style roles (\textit{Agent}, \textit{Location}, \textit{Manner}) concentrate in early layers (L0--L1); \textit{Time} neurons appear in every layer; \textit{Quantity} neurons dominate the top-20 cap throughout the network.}
\label{fig:neuron-selectivity}
\end{figure}

\subsection{Co-activation Correlation Methodology}
\label{app:coactivation-method}

The co-activation correlations ($\rho$) reported in Section~\ref{sec:neurons} are computed as follows:

\begin{enumerate}
    \item \textbf{Collect activations}: For each QA-SRL validation sample, we extract the feed-forward layer activations $\mathbf{a} \in \mathbb{R}^{d_{\text{ff}}}$ at each layer.
    \item \textbf{Filter to role-selective neurons}: We retain only neurons identified as role-selective via PCA (see Section~\ref{sec:neurons}), yielding activation vectors $\mathbf{a}_{\text{selective}} \in \mathbb{R}^{n_{\text{selective}}}$.
    \item \textbf{Compute pairwise Pearson correlations}: For each pair of role-selective neurons $(i, j)$, we compute the Pearson correlation coefficient across all $N$ validation samples:
    \begin{equation}
    \rho_{ij} = \frac{\sum_{k=1}^{N}(a_i^{(k)} - \bar{a}_i)(a_j^{(k)} - \bar{a}_j)}{\sqrt{\sum_{k=1}^{N}(a_i^{(k)} - \bar{a}_i)^2}\sqrt{\sum_{k=1}^{N}(a_j^{(k)} - \bar{a}_j)^2}}
    \end{equation}
    where $a_i^{(k)}$ is the activation of neuron $i$ on sample $k$.
    \item \textbf{Partition by role relationship}: We separate correlations into two groups based on whether the neuron pair shares the same semantic role:
    \begin{align}
    \rho_{\text{within}} &= \{\rho_{ij} : \text{role}(i) = \text{role}(j)\} \\
    \rho_{\text{across}} &= \{\rho_{ij} : \text{role}(i) \neq \text{role}(j)\}
    \end{align}
    \item \textbf{Aggregate}: We report the mean within-role correlation ($\bar{\rho}_{\text{within}} \approx 0.24$) and mean across-role correlation ($\bar{\rho}_{\text{across}} \approx 0.10$). The difference $\Delta\rho = \bar{\rho}_{\text{within}} - \bar{\rho}_{\text{across}}$ measures functional grouping strength.
\end{enumerate}

\begin{figure}[h]
\centering
\includegraphics[width=\linewidth]{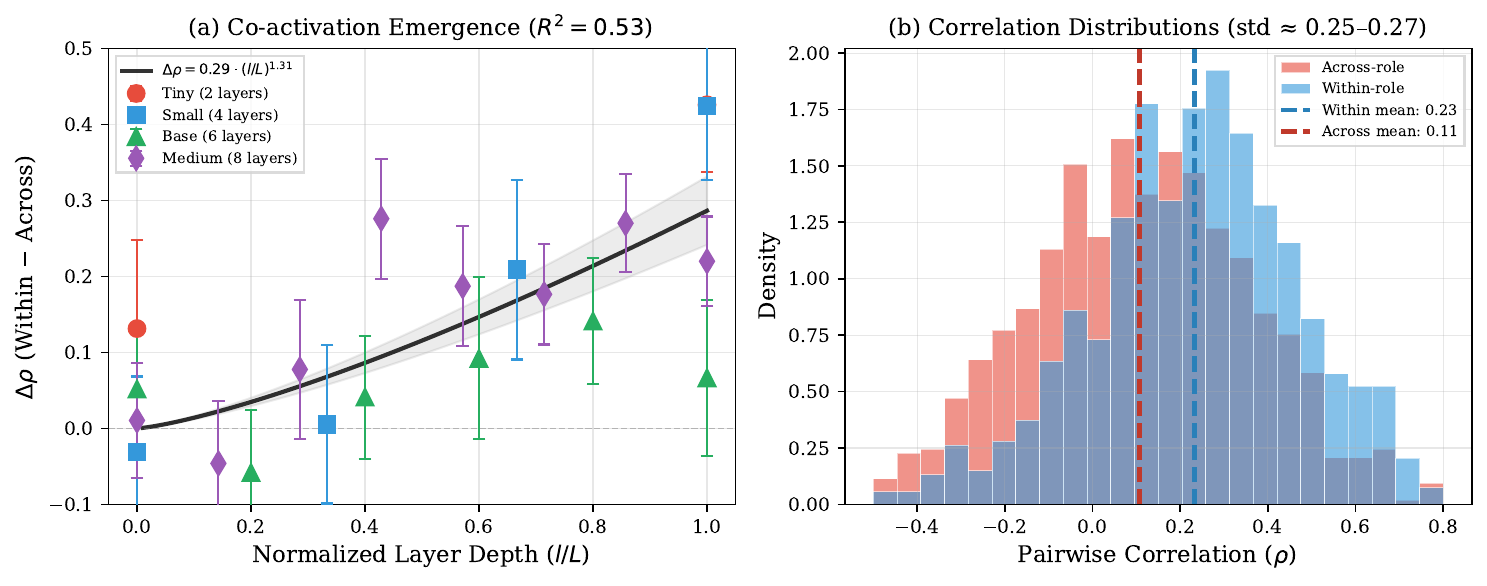}
\caption{Role-specific neuron separation is weak but increases with depth. \textit{Top:} Each point shows $\Delta\rho = \rho_{\text{within}} - \rho_{\text{across}}$ for one layer, measuring whether same-role neurons co-activate more than different-role neurons. Error bars show approximate standard errors. We fit a power-law $\Delta\rho(l) = \alpha \cdot (l/L)^\beta$, obtaining $\alpha = 0.29 \pm 0.05$, $\beta = 1.31 \pm 0.51$, $R^2 = 0.53$. \textit{Bottom:} The underlying distributions of within-role (blue) and across-role (red) correlations. The distributions heavily overlap (std $\approx$ 0.3--0.4 vs.\ mean differences of 0.1--0.2), indicating that role-specific groupings are weak but present.}
\label{fig:powerlaw_fit}
\end{figure}

\section{Semantic Role Representations Across Scales}
\label{app:representations}

We visualize PropBank-style semantic role representations for all model scales using PCA and t-SNE.

\begin{figure}[h]
\centering
\begin{subfigure}[b]{0.48\textwidth}
\includegraphics[width=\textwidth]{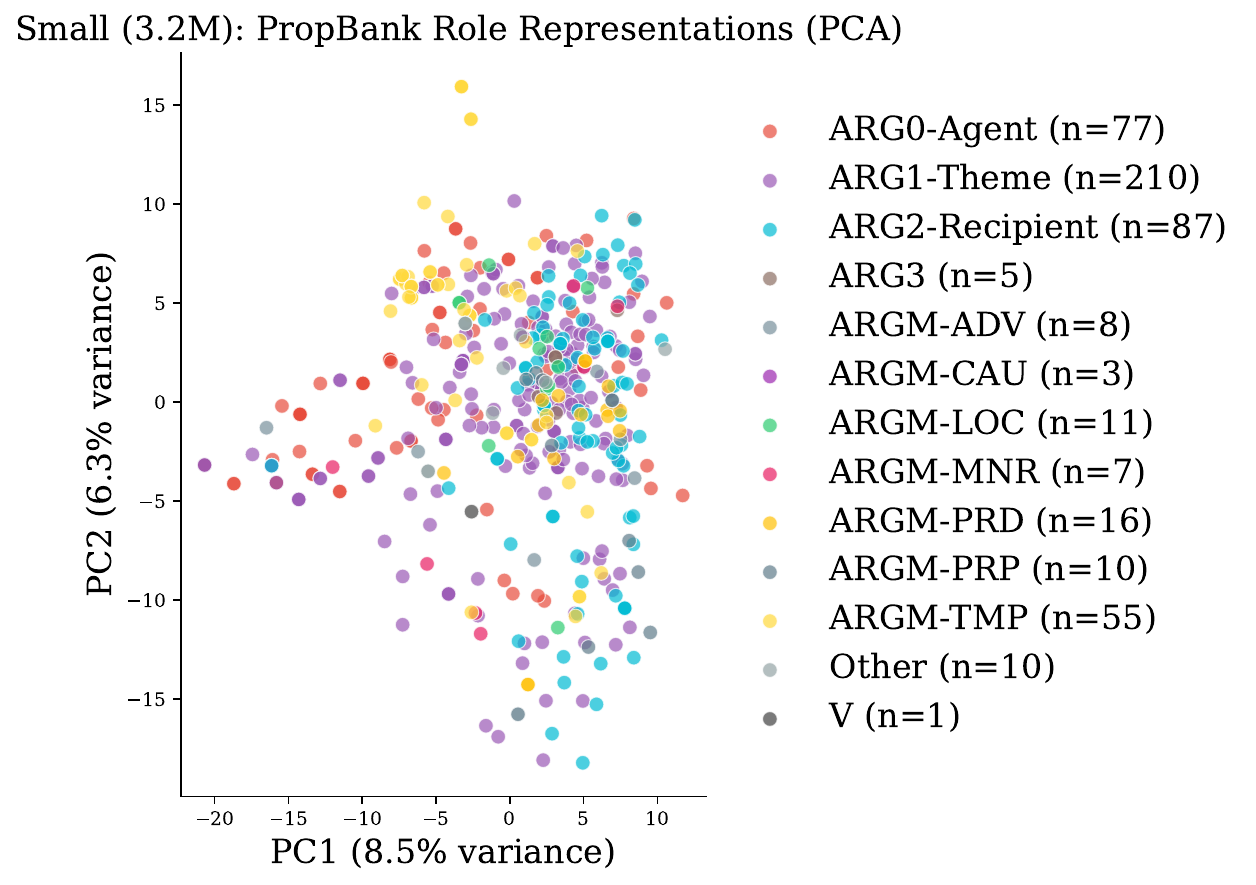}
\caption{PCA visualization}
\end{subfigure}
\hfill
\begin{subfigure}[b]{0.48\textwidth}
\includegraphics[width=\textwidth]{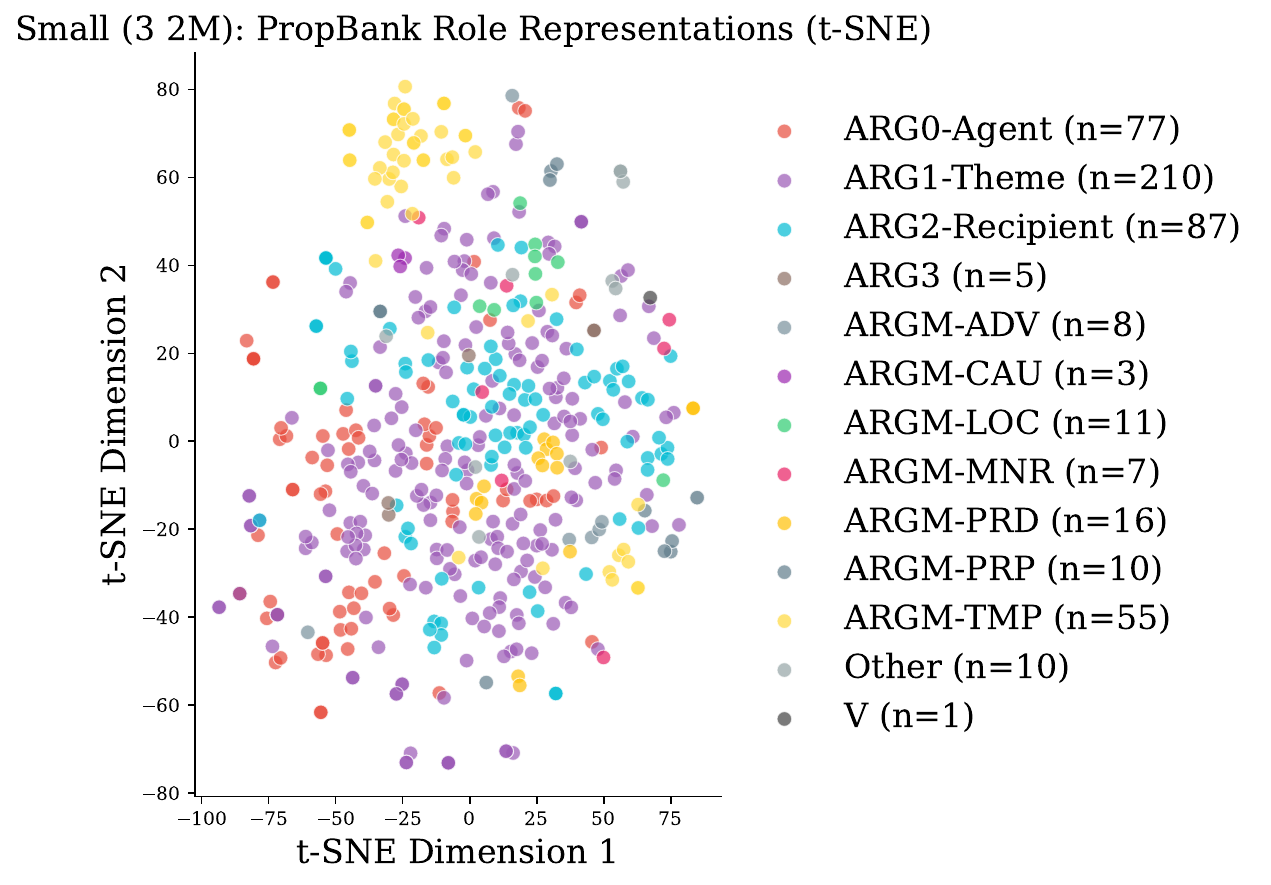}
\caption{t-SNE visualization}
\end{subfigure}
\caption{Small model (3.2M transformer parameters): PropBank role representations. Final layer activations colored by PropBank role; ARG0-Agent shows partial clustering, while ARG1 variants and adjuncts are distributed across the space. PCA explains 14.8\% of variance in the first two components.}
\label{fig:role-representations}
\end{figure}

\begin{figure}[h]
\centering
\begin{subfigure}[b]{0.48\textwidth}
\includegraphics[width=\textwidth]{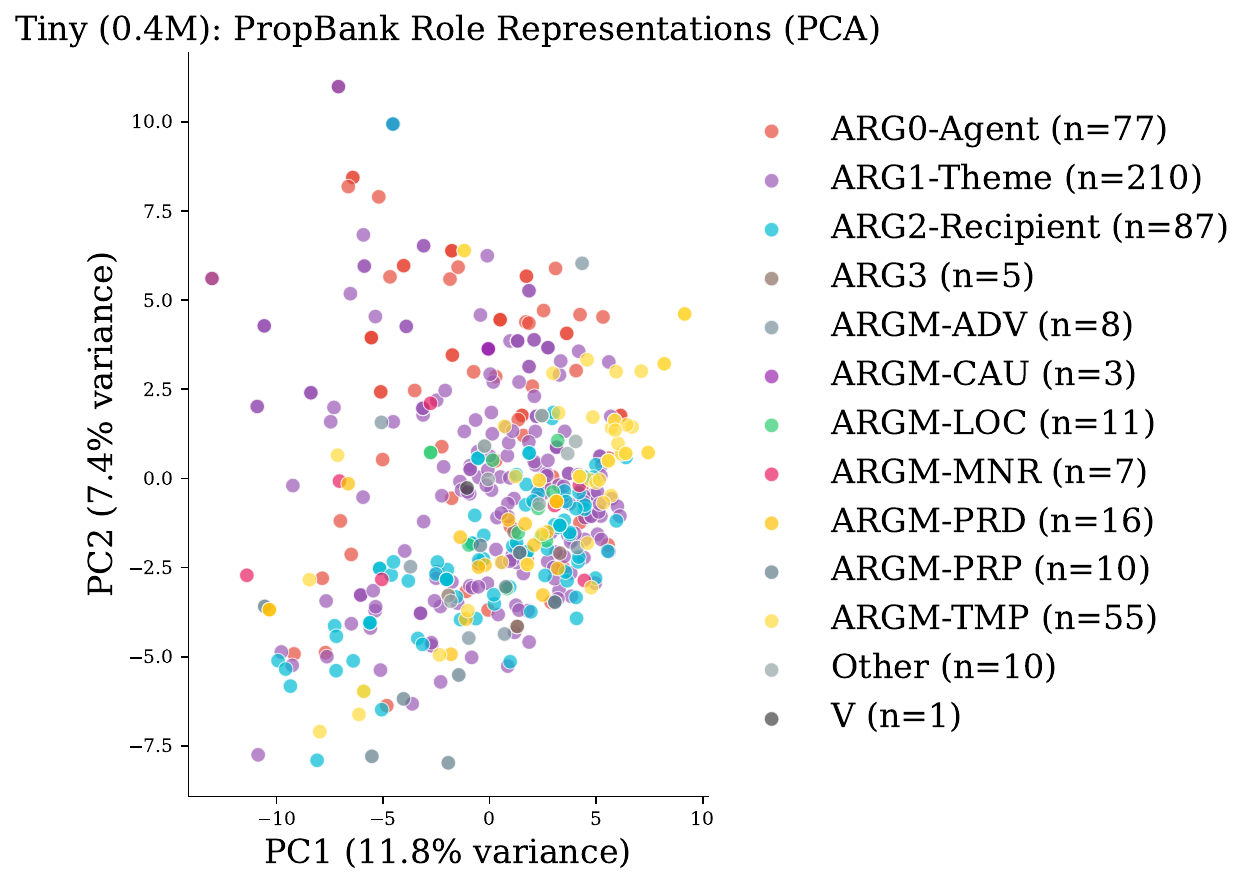}
\caption{PCA visualization.}
\end{subfigure}
\hfill
\begin{subfigure}[b]{0.48\textwidth}
\includegraphics[width=\textwidth]{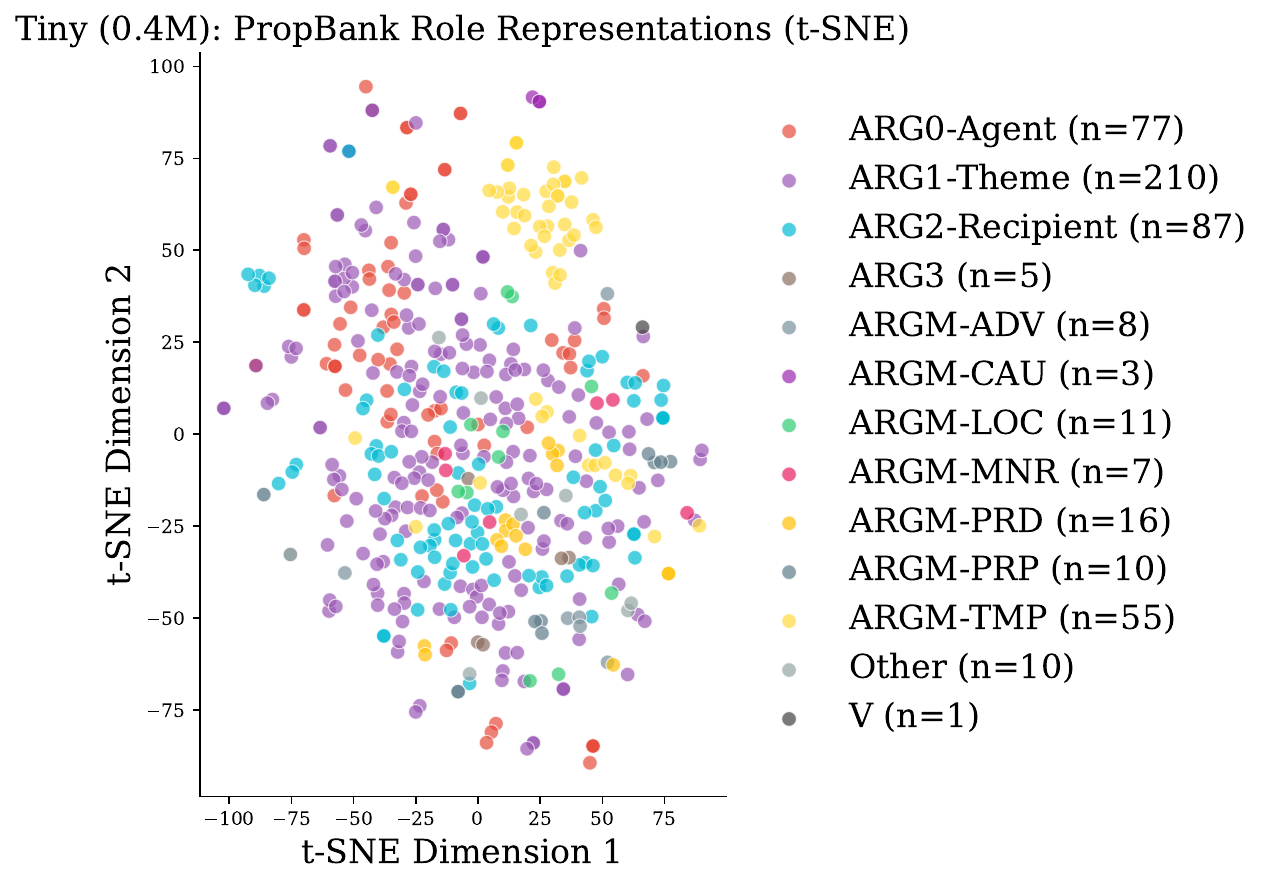}
\caption{t-SNE visualization.}
\end{subfigure}
\caption{Tiny model (0.4M transformer parameters): PropBank role representations. PCA (left) and t-SNE (right) visualization of final layer activations colored by PropBank role. Even at this minimal scale, ARG0-Agent shows some separation. PCA explains 19.2\% of variance in first two components, more than larger models, suggesting simpler/more concentrated representations.}
\label{fig:tiny-representations}
\end{figure}

\begin{figure}[h]
\centering
\begin{subfigure}[b]{0.48\textwidth}
\includegraphics[width=\textwidth]{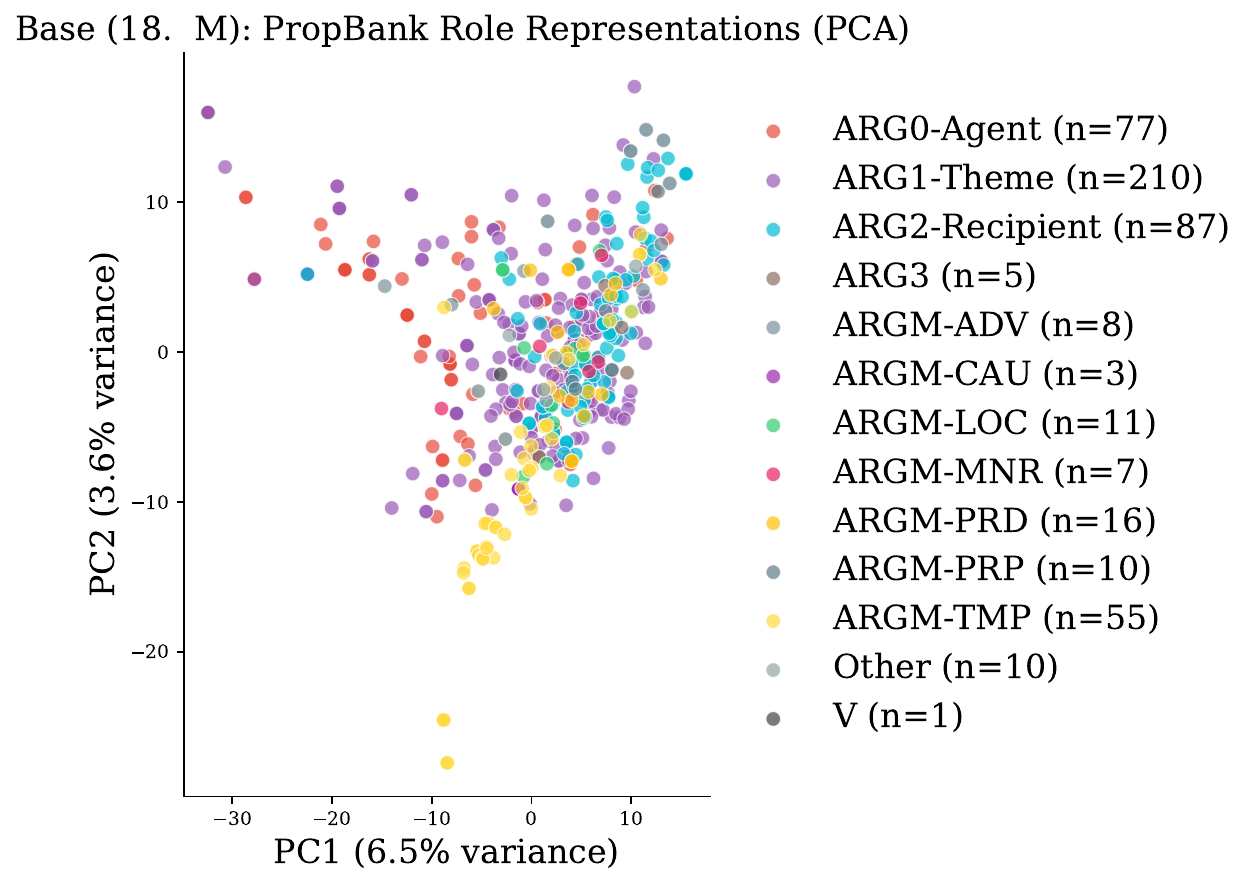}
\caption{PCA visualization.}
\end{subfigure}
\hfill
\begin{subfigure}[b]{0.48\textwidth}
\includegraphics[width=\textwidth]{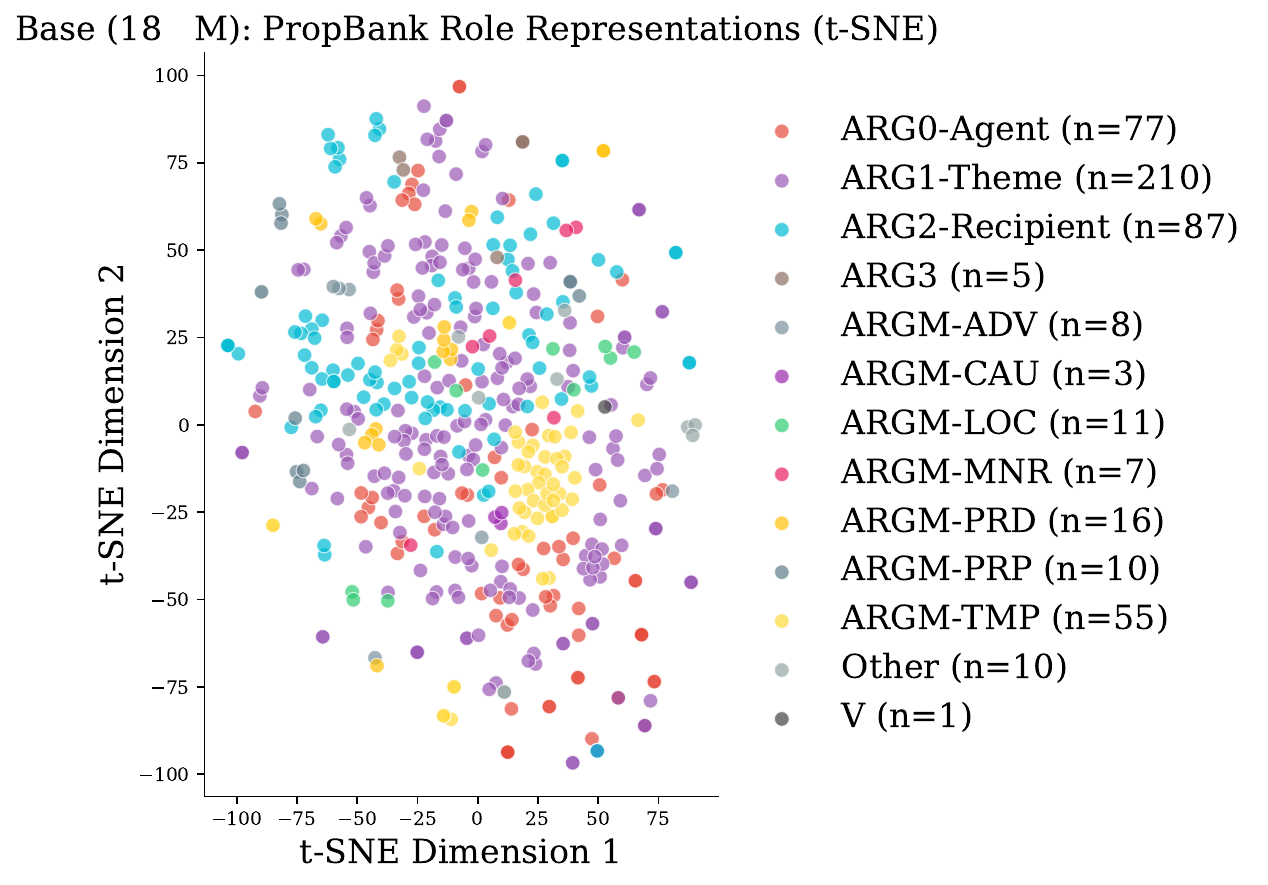}
\caption{t-SNE visualization.}
\end{subfigure}
\caption{Base model (18.9M transformer parameters): PropBank role representations. PCA (left) and t-SNE (right) visualization of final layer activations colored by PropBank role. Similar to the Small model, ARG0-Agent shows partial clustering, while other roles (ARG1 variants, ARGM adjuncts) overlap substantially. PCA explains 10.1\% of variance in first two components.}
\label{fig:base-representations}
\end{figure}

\begin{figure}[h]
\centering
\begin{subfigure}[b]{0.48\textwidth}
\includegraphics[width=\textwidth]{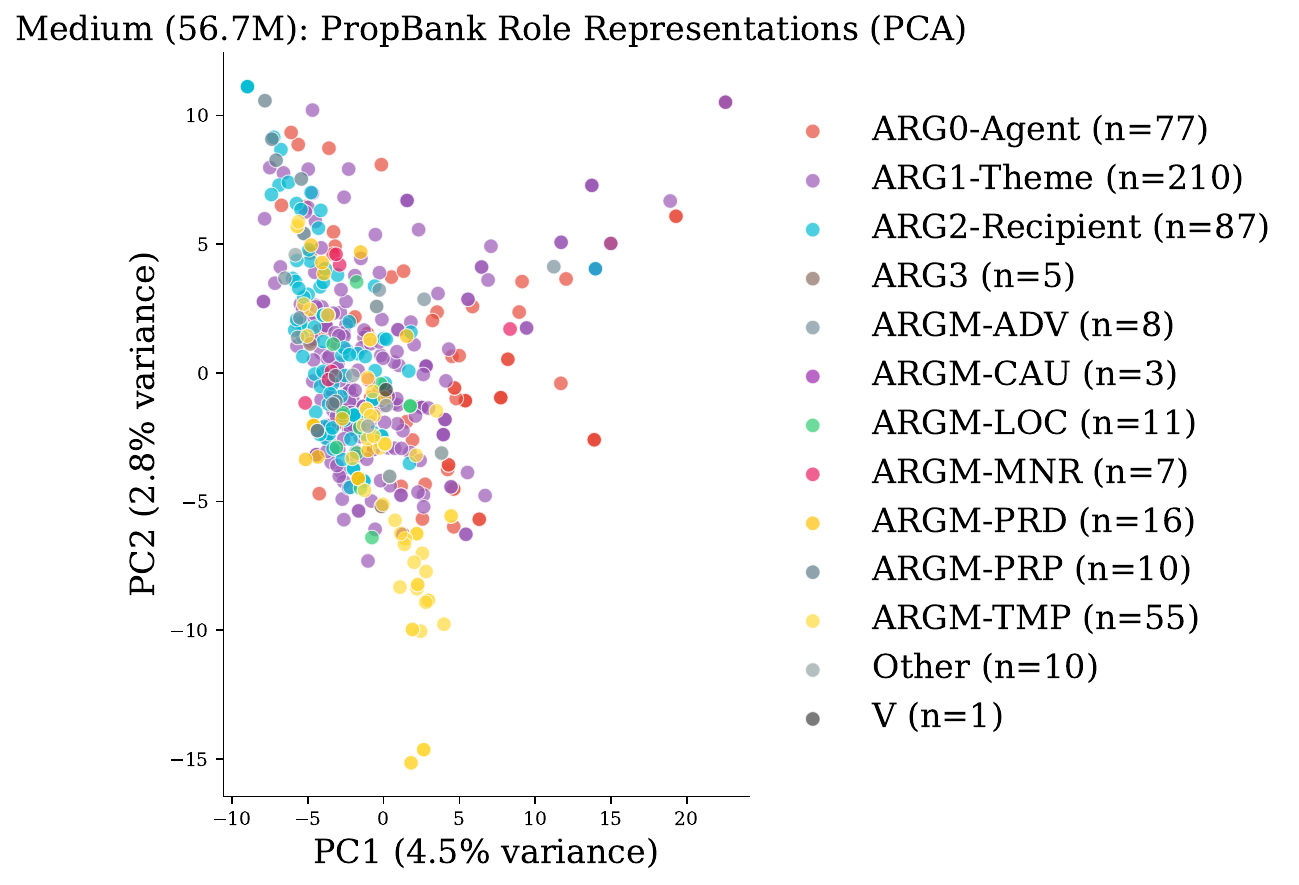}
\caption{PCA visualization.}
\end{subfigure}
\hfill
\begin{subfigure}[b]{0.48\textwidth}
\includegraphics[width=\textwidth]{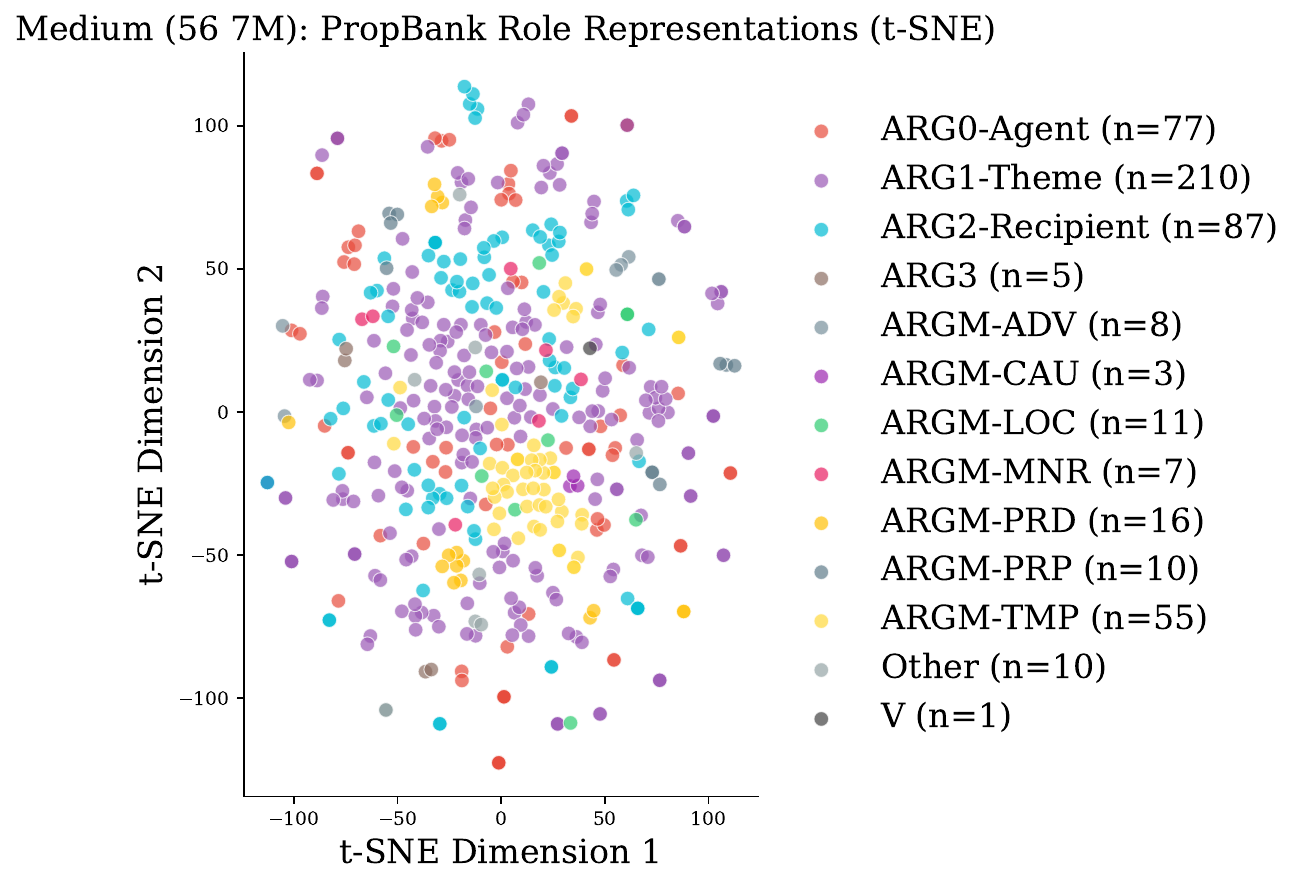}
\caption{t-SNE visualization.}
\end{subfigure}
\caption{Medium model (56.7M transformer parameters): PropBank role representations. PCA (left) and t-SNE (right) visualization of final layer activations colored by PropBank role. The Medium model shows similar patterns to smaller scales, with ARG0-Agent partially separated but most roles distributed across representation space. PCA explains only 7.3\% of variance, suggesting representations become more distributed with scale.}
\label{fig:medium-representations}
\end{figure}

\section{PropBank Label Assignment}
\label{app:propbank}

The QA-SRL dataset provides question-answer pairs but not explicit PropBank semantic role labels. To obtain PropBank labels for our analysis, we use AllenNLP's BERT-based SRL model \citep{shi2019simple}, specifically the \texttt{structured-prediction-srl-bert.2020.12.15} checkpoint trained on OntoNotes 5.0. This model achieves 86.5 F1 on the CoNLL-2012 SRL benchmark.

\paragraph{Labeling Process.} For each QA-SRL example (sentence, question, answer), we:
\begin{enumerate}
    \item Run the AllenNLP SRL predictor on the sentence to obtain PropBank argument structure for each verb.
    \item Match the QA-SRL answer span to predicted SRL argument spans using character overlap.
    \item Assign the PropBank label if $\geq$50\% of the answer overlaps with an SRL argument; otherwise mark as ``Unknown.''
\end{enumerate}

\paragraph{Label Categories.} For role-conditioned analyses (correlation, neuron-ablation $\Delta$F1) we collapse PropBank labels into the following readable categories:
\begin{itemize}
    \item \textbf{ARG0-Agent}: Proto-agent (doer, causer).
    \item \textbf{ARG1-Theme}: Proto-patient (undergoer, theme).
    \item \textbf{ARG2-Other}: Varies by verb (beneficiary, instrument, attribute, recipient, quantity).
    \item \textbf{ARGM-LOC}: Location.
    \item \textbf{ARGM-TMP}: Temporal/Time.
    \item \textbf{ARGM-MNR}: Manner (how the action was performed).
    \item \textbf{ARGM-CAU}: Cause.
    \item \textbf{ARGM-DIR}: Direction.
\end{itemize}

\paragraph{Raw labels in visualizations.} The PCA and t-SNE scatter plots in Appendix~\ref{app:representations} (Figures~\ref{fig:role-representations}--\ref{fig:medium-representations}) and the role-selectivity heatmap in Figure~\ref{fig:neuron-selectivity} display the full set of raw labels emitted by AllenNLP rather than the collapsed categories above, in order to expose any sub-role structure to the reader. The additional labels visible in those figures (\texttt{ARG2-Recipient}, \texttt{ARG3}, \texttt{ARGM-ADV}, \texttt{ARGM-PRD}, \texttt{ARGM-PRP}, and the \texttt{Quantity} category in Figure~\ref{fig:neuron-selectivity}) are AllenNLP's verb-specific or adverbial sub-roles. They are subsumed under \textbf{ARG2-Other} or treated as out-of-vocabulary noise in the collapsed scheme used for all numerical analyses, so the figures and tables refer to slightly different label inventories by design.

\section{Replication on GPT-2}
\label{app:gpt2}

To rule out that the Agent-neuron sign reversal is an artifact of our specific training setup, we replicate the causal neuron ablation protocol on OpenAI's GPT-2 \citep{radford2019language} at two scales: GPT-2 Small (124M parameters, 12 layers) and GPT-2 Medium (355M parameters, 24 layers). GPT-2 was trained on WebText, providing approximately $80\times$ more tokens per parameter than our custom Medium model and using a different corpus.

\paragraph{Protocol difference: no-retrain ablation.} Custom models use \textit{retrain} ablation: weights are zeroed in the pre-trained checkpoint, then the model is fine-tuned on QA-SRL and evaluated. For GPT-2 and Pythia we instead use \textit{no-retrain} ablation: we freeze the pre-trained backbone, train only a QA head (matching the frozen probing protocol), zero the targeted neurons in the resulting probe, and re-evaluate without further training. This is necessary because we do not fine-tune off-the-shelf GPT-2/Pythia in the main protocol (the goal is to probe pre-training, not adapt these large models), and re-running fine-tuning at every ablation site would multiply the compute by orders of magnitude. The two protocols measure related but distinct causal properties: retrain ablation tests whether a neuron is necessary for \textit{learning} the role, while no-retrain ablation tests whether it is necessary for \textit{using} role information already present after probing. We report both signs of the Agent effect across families because the central claim is qualitative (sign attenuation as a function of scale), not a quantitative comparison of $\Delta$F1 magnitudes between families. Random-neuron controls (third column of Tables~\ref{tab:gpt2-ablation} and~\ref{tab:pythia-ablation}) show near-zero effects under the no-retrain protocol, ruling out trivial perturbation artifacts.

\begin{table}[h]
\caption{GPT-2 neuron ablation results (5 seeds each). Agent-neurons are strongly causal in GPT-2 Small ($-9.4\%$) and negligible in GPT-2 Medium ($-0.8\%$), matching the sign-attenuation pattern in our custom models. Random neuron controls show near-zero effects.}
\label{tab:gpt2-ablation}
\begin{center}
\begin{tabular}{llcc}
\toprule
Model & Role & No-retrain $\Delta$ & Random control $\Delta$ \\
\midrule
GPT-2 Small (124M) & Agent & $-9.4 \pm 2.1\%$ & $+0.1 \pm 0.8\%$ \\
GPT-2 Small (124M) & Time & $-0.5 \pm 0.7\%$ & $0.0 \pm 0.5\%$ \\
GPT-2 Medium (355M) & Agent & $-0.8 \pm 0.7\%$ & $0.0 \pm 0.5\%$ \\
GPT-2 Medium (355M) & Time & $-3.9 \pm 1.6\%$ & $+0.4 \pm 0.6\%$ \\
\bottomrule
\end{tabular}
\end{center}
\end{table}

Layer-wise probing on GPT-2 confirms a depth gradient consistent with Section~\ref{sec:layerwise}: the best probing layer is at 92\% depth in GPT-2 Small (layer 11, F1 = 31.2\%) and 71\% depth in GPT-2 Medium (layer 17, F1 = 35.1\%).

\begin{figure}[h]
\centering
\includegraphics[width=0.7\linewidth]{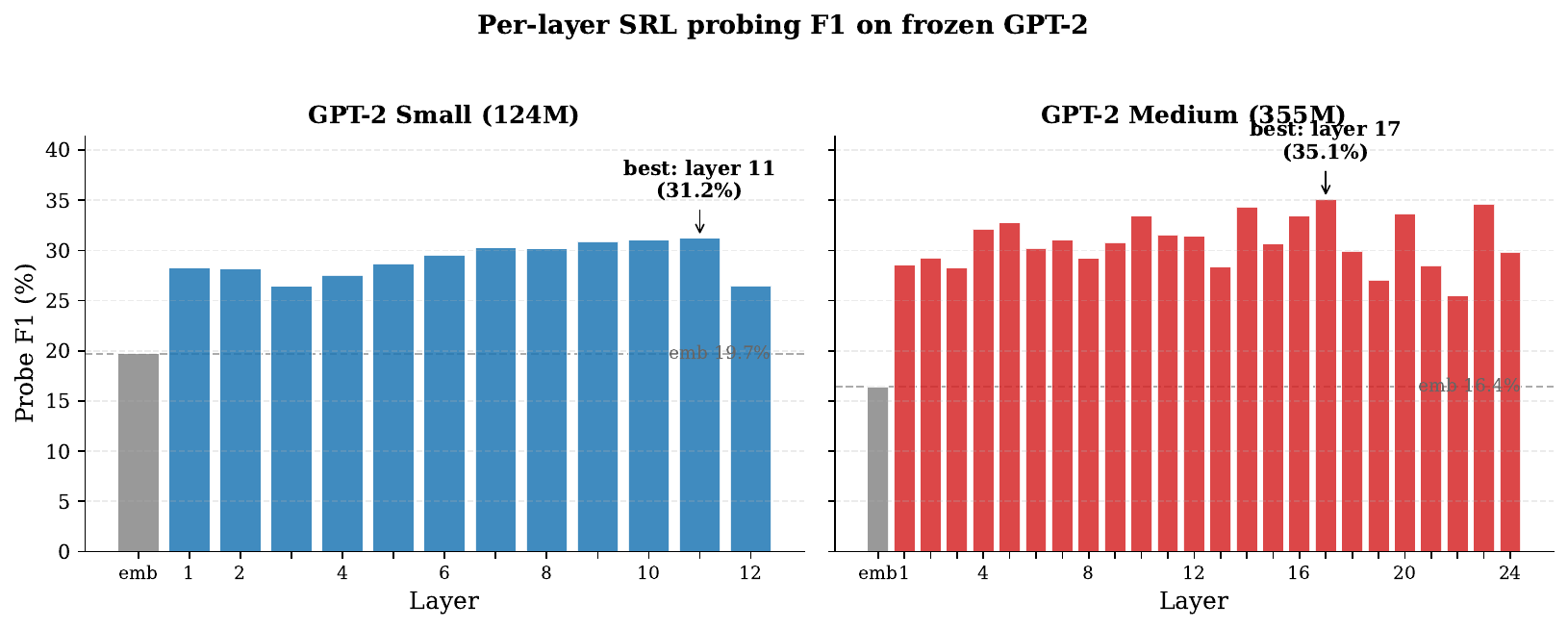}
\caption{Layer-wise probing F1 on GPT-2 Small and Medium. SRL information peaks in deeper layers, with the best layer at 92\% depth (Small) and 71\% depth (Medium).}
\label{fig:gpt2-probing}
\end{figure}

\section{Replication on Pythia}
\label{app:pythia}

We further replicate on EleutherAI's Pythia \citep{biderman2023pythia}, a third model family trained on The Pile \citep{gao2020pile}, a different corpus from both WikiText-103 and WebText. Pythia provides three additional scales (70M, 160M, 410M).

\begin{table}[h]
\caption{Pythia neuron ablation results (5 seeds each). Agent neuron causal effects attenuate from $-8.3\%$ at 70M to $+0.5\%$ at 410M, consistent with the sign-attenuation pattern in our custom and GPT-2 results. Random controls confirm role-specificity.}
\label{tab:pythia-ablation}
\begin{center}
\begin{tabular}{llcc}
\toprule
Model & Role & No-retrain $\Delta$ & Random control $\Delta$ \\
\midrule
Pythia-70M & Agent & $-8.3 \pm 1.3\%$ & $-0.4 \pm 0.8\%$ \\
Pythia-160M & Agent & $-1.5 \pm 1.1\%$ & $+0.2 \pm 0.5\%$ \\
Pythia-410M & Agent & $+0.5 \pm 1.5\%$ & $+0.1 \pm 0.6\%$ \\
Pythia-410M & Time & $-1.2 \pm 0.9\%$ & $+0.2 \pm 0.3\%$ \\
\bottomrule
\end{tabular}
\end{center}
\end{table}

The depth gradient also holds: the best probing layer is at 17\% depth (Pythia-70M), 33\% depth (Pythia-160M), and 42\% depth (Pythia-410M).

\section{Component-Level Circuit Analysis}
\label{app:circuit}

To characterize the transition from concentrated to distributed organization, we zero-ablated each attention layer and MLP layer individually (3 seeds each) across all six models and measured the causal effect on Agent F1.

\begin{table}[h]
\caption{Component-level circuit analysis. The number of components (attention + MLP layers) needed for 80\% of total causal importance increases with scale, while the concentration ratio decreases.}
\label{tab:circuit-analysis}
\begin{center}
\begin{tabular}{lccc}
\toprule
Model & Components for 80\% & Total components & Concentration ratio \\
\midrule
Custom Tiny (0.4M) & 2.3 & 4 & 58\% \\
Custom Small (3.2M) & 4.7 & 8 & 58\% \\
Custom Base (18.9M) & 6.3 & 12 & 53\% \\
Custom Medium (57M) & 7.3 & 16 & 46\% \\
GPT-2 Small (124M) & 11.3 & 24 & 47\% \\
GPT-2 Medium (355M) & 21.3 & 48 & 44\% \\
\bottomrule
\end{tabular}
\end{center}
\end{table}

\section{Emergence During Pre-training}
\label{app:pretraining-emergence}

We probed intermediate pre-training checkpoints (every 5th epoch) across all four custom scales to track when SRL information appears.

\begin{figure}[h]
\centering
\includegraphics[width=0.95\linewidth]{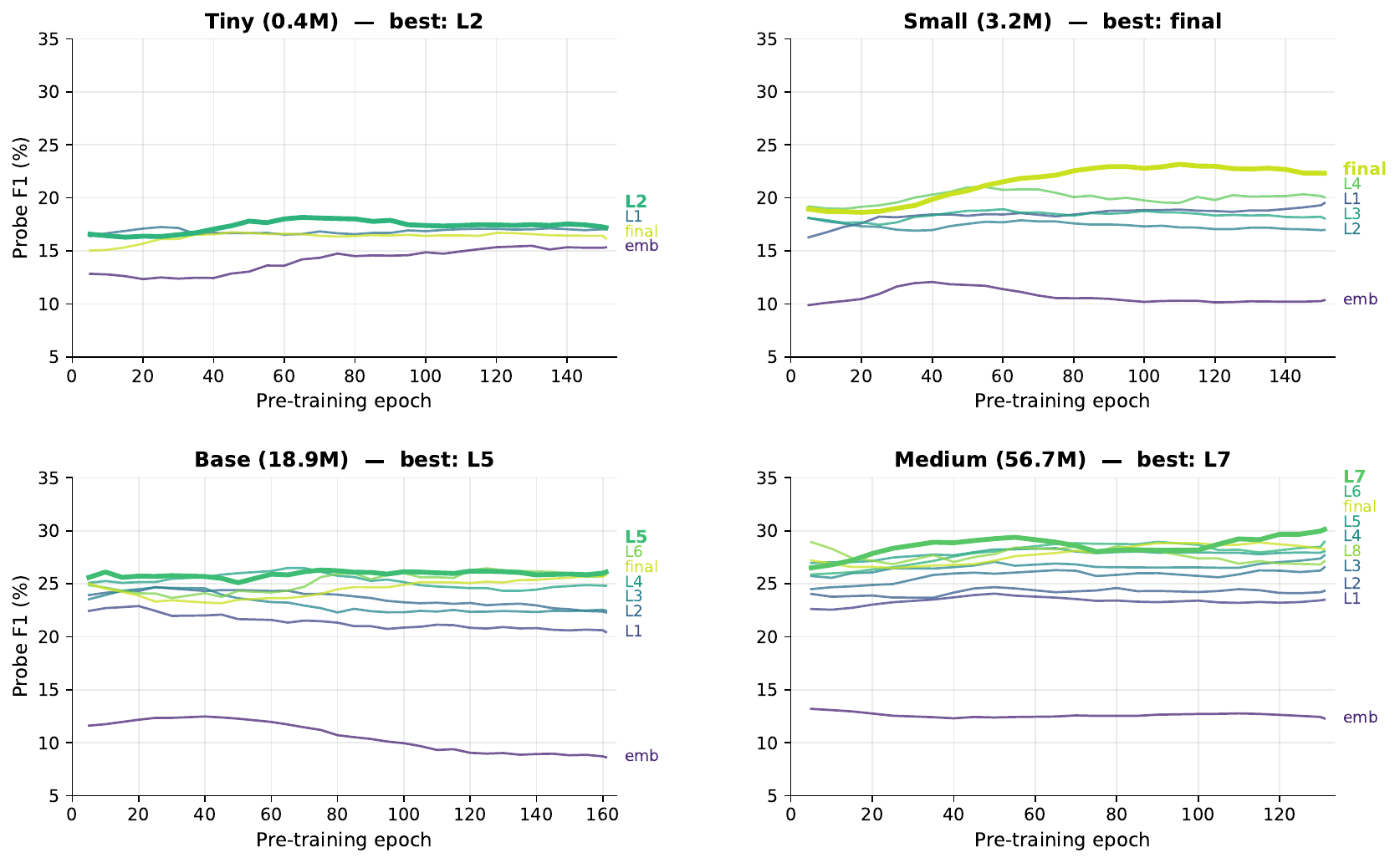}
\caption{Per-layer probe F1 across pre-training, one panel per scale. Each line is one layer (embeddings, transformer blocks, final norm), labelled at its right endpoint; the best layer per scale is bolded and named in the panel title. F1 stabilises within the first 10--20 epochs and remains roughly flat thereafter, with embeddings persistently $\sim$10 F1 below transformer-block layers. Across panels, the best layer deepens with scale (L2 in Tiny, final in Small, L5 in Base, L7 in Medium), and steady-state F1 grows monotonically with scale ($\sim$17\% / $\sim$22\% / $\sim$26\% / $\sim$30\%).}
\label{fig:pretraining-emergence}
\end{figure}

\begin{table}[h]
\caption{Pre-training convergence status. The Medium model has not converged at its final checkpoint, suggesting our emergence scores are conservative lower bounds.}
\label{tab:pretraining-convergence}
\begin{center}
\begin{tabular}{llcl}
\toprule
Model & Final best layer & Probe F1 & Status \\
\midrule
Tiny (0.4M) & block\_1 & 17.2\% & Stable from early training \\
Small (3.2M) & final\_norm & 22.5\% & Slowly rising \\
Base (18.9M) & block\_5 & 26.3\% & Stable by mid-training \\
Medium (57M) & block\_7 & 30.8\% & Not converged at epoch 131 \\
\bottomrule
\end{tabular}
\end{center}
\end{table}

\section{Frozen-Embedding (SEP-only) Ablation}
\label{app:sep-only}

To isolate what is encoded in pre-trained transformer weights, we freeze the entire token embedding matrix except \texttt{[SEP]} (kept trainable so the model can disambiguate question from sentence), alongside positional embeddings and transformer blocks; only \texttt{[SEP]} and the QA head update.

\begin{table}[h]
\caption{Frozen-embedding (SEP-only) ablation, 5 seeds. With the token embedding matrix held fixed except \texttt{[SEP]}, the residual transformer contribution grows monotonically with scale and is significant at Base ($p=0.004$) and Medium ($p=0.002$); Tiny/Small are non-significant, reflecting a capacity bottleneck on routing task information through a single trainable token across only 2--4 blocks.}
\label{tab:sep-only}
\centering
\small
\begin{tabular}{lccl}
\toprule
Scale & Pretrained F1 & Random F1 & Emergence \\
\midrule
Tiny   & $23.9 \pm 1.0$ & $24.4 \pm 1.3$ & $-0.5 \pm 1.1\%$ (ns, $p=0.38$) \\
Small  & $27.9 \pm 1.1$ & $27.4 \pm 0.7$ & $+0.5 \pm 1.9\%$ (ns, $p=0.59$) \\
Base   & $30.5 \pm 0.8$ & $27.8 \pm 0.4$ & $+2.7 \pm 1.0\%$ ($t=6.2$, $p=0.004$) \\
Medium & $31.6 \pm 0.4$ & $28.0 \pm 0.7$ & $+3.6 \pm 1.1\%$ ($t=7.5$, $p=0.002$) \\
\bottomrule
\end{tabular}
\end{table}

The residual 3--4 F1 gap at Base/Medium is smaller than the 12--19 F1 frozen-probe gap, showing the original probe condition combines transformer-encoded emergence with embedding adaptation. Combined with the zero-adaptation per-layer probes (Section~\ref{sec:layerwise}, 10--26\% F1), SEP-only establishes that SRL emergence in transformer weights is statistically significant at Base/Medium; at Tiny/Small the single-token bottleneck limits what the SEP-only setup can detect, while the per-layer probes still recover above-chance F1 at those scales.


\end{document}